\documentclass[10pt,twocolumn,letterpaper]{article}

\usepackage{cvpr}
\usepackage{times}
\usepackage{epsfig}
\usepackage{graphicx}
\usepackage{amsmath}
\usepackage{amssymb}

\usepackage{blindtext}%
\usepackage{caption}
\usepackage{booktabs}
\usepackage{multirow}
\usepackage{mathtools}
\usepackage{bm}
\usepackage{array}
\makeatletter
\@namedef{ver@everyshi.sty}{}
\makeatother
\usepackage[mode=buildnew,subpreambles]{standalone}
\usepackage{tikz}

\newcolumntype{P}[1]{>{\centering\arraybackslash}p{#1}}

\usepackage[pagebackref=true,breaklinks=true,letterpaper=true,colorlinks,bookmarks=false]{hyperref}

\cvprfinalcopy 

\def\cvprPaperID{9142} 

\ifcvprfinal\fi

\newcommand\copyrighttext{%
	\footnotesize \textcopyright 2020 IEEE. Personal use of this material is permitted.
	Permission from IEEE must be obtained for all other uses, in any current or future
	media, including reprinting/republishing this material for advertising or promotional
	purposes, creating new collective works, for resale or redistribution to servers or
	lists, or reuse of any copyrighted component of this work in other works.
	Presented at the 2020 IEEE/CVF Conference on Computer Vision and Pattern Recognition (CVPR).
	DOI: 10.1109/CVPR42600.2020.00961.
	Publisher version: https://ieeexplore.ieee.org/document/9156288.}
\newcommand\copyrightnotice{%
	\begin{tikzpicture}[remember picture,overlay]
	\node[anchor=south west,yshift=10pt,xshift=1.6cm] at (current page.south west) {\parbox{\textwidth}{\copyrighttext}};
	\end{tikzpicture}%
}

\begin{document}

\title{Where Does It End? -- Reasoning About Hidden Surfaces by Object Intersection Constraints}

\author{Michael Strecke and J{\"o}rg St{\"u}ckler\\
Embodied Vision Group, Max Planck Institute for Intelligent Systems, T{\"u}bingen\\
{\tt\small \{mstrecke,jstueckler\}@tue.mpg.de}
}

\maketitle
\copyrightnotice
\thispagestyle{empty}

\begin{abstract}
Dynamic scene understanding is an essential capability in robotics and VR/AR.
In this paper we propose Co-Section, an optimization-based approach to 3D dynamic scene reconstruction, which infers hidden shape information from intersection constraints.
An object-level dynamic SLAM frontend detects, segments, tracks and maps dynamic objects in the scene.
Our optimization backend completes the shapes using hull and intersection constraints between the objects.
In experiments, we demonstrate our approach on real and synthetic dynamic scene datasets.
We also assess the shape completion performance of our method quantitatively.
To the best of our knowledge, our approach is the first method to incorporate such physical plausibility constraints on object intersections for shape completion of dynamic objects in an energy minimization framework.
\end{abstract}

\section{Introduction}
\begin{figure}
	\centering
	\footnotesize
	\setlength{\tabcolsep}{1pt}
	\renewcommand{\arraystretch}{0.6}
	\begin{tabular}{P{.32\linewidth}P{.32\linewidth}P{.32\linewidth}}
		\includegraphics[width=\linewidth]{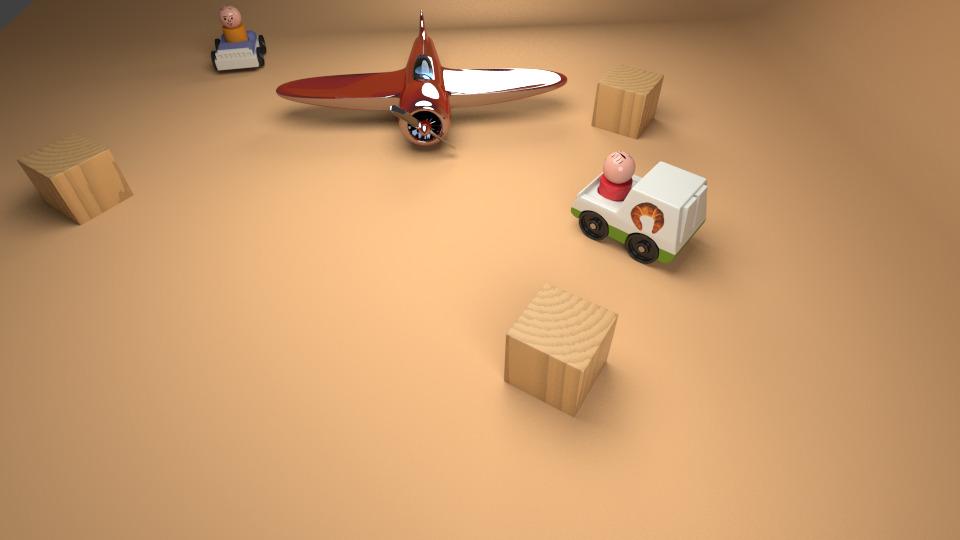} & \includegraphics[width=\linewidth]{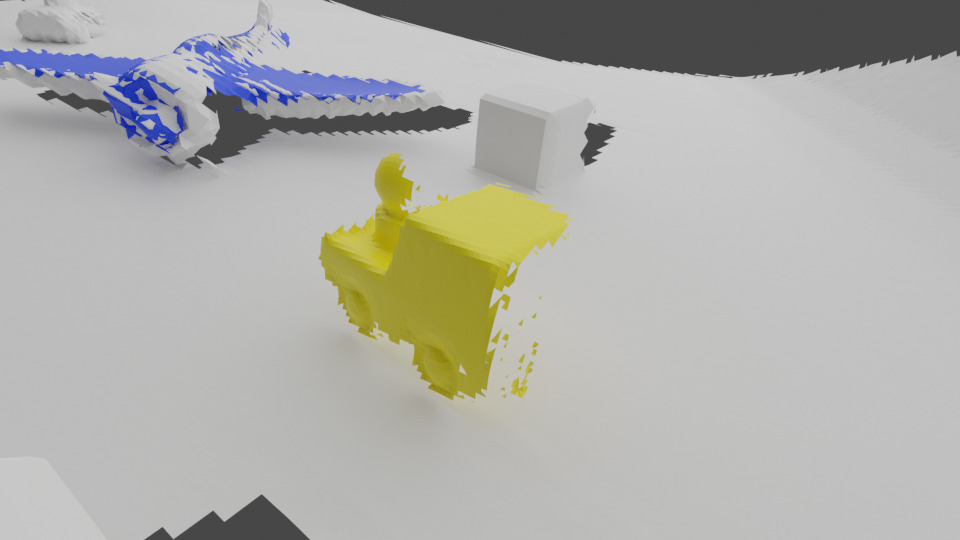} & \includegraphics[width=\linewidth]{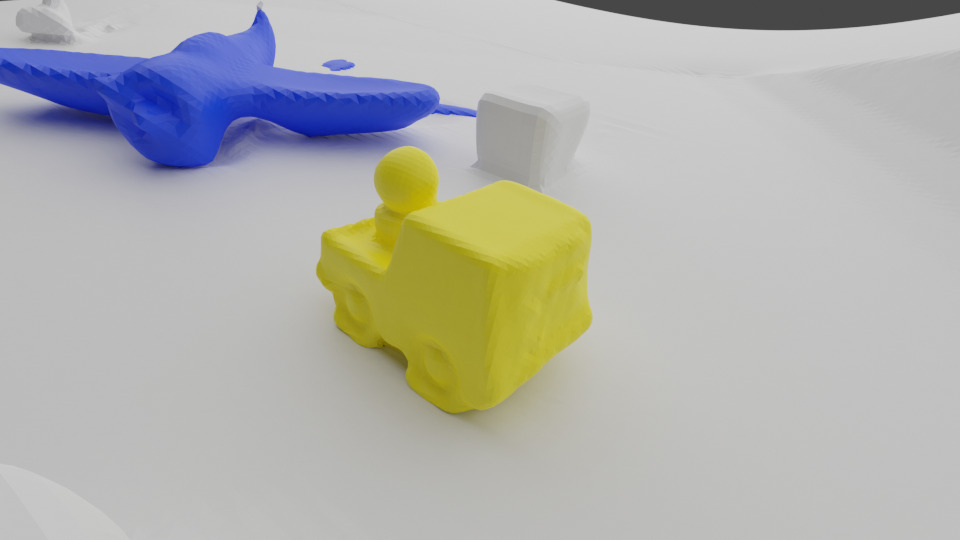} \\
		&  \includegraphics[width=\linewidth]{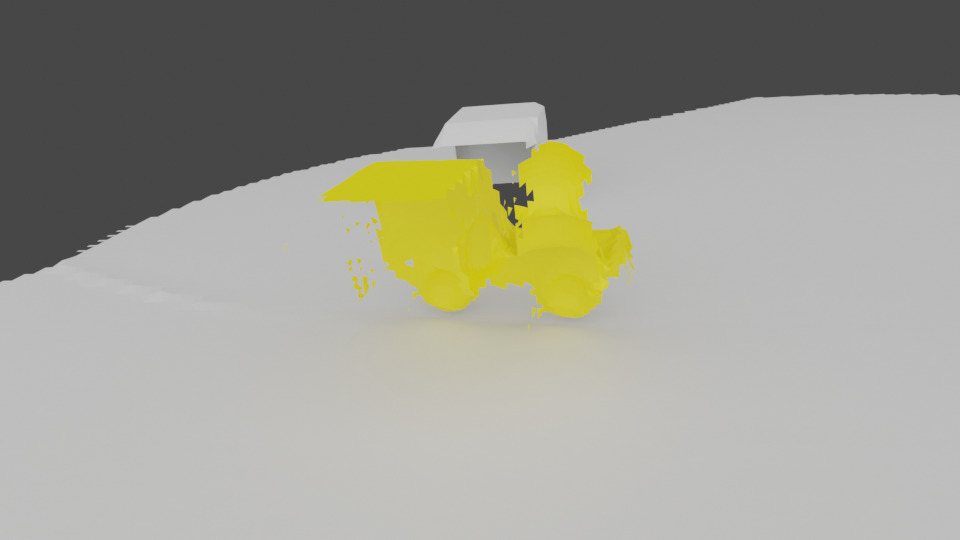} & \includegraphics[width=\linewidth]{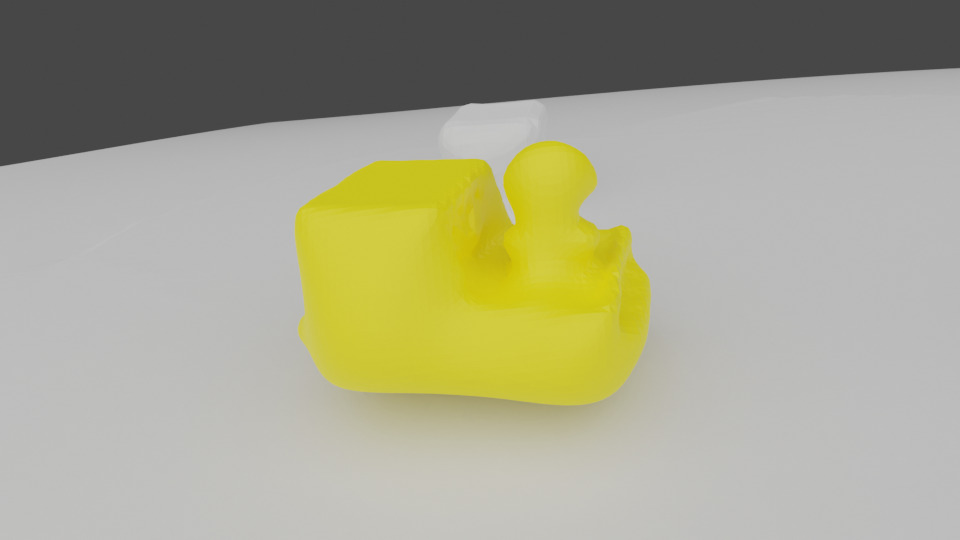} \\
		\includegraphics[width=\linewidth]{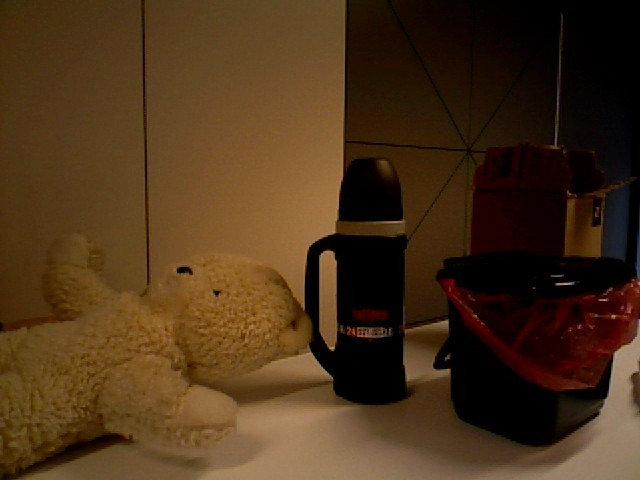} & \includegraphics[width=\linewidth]{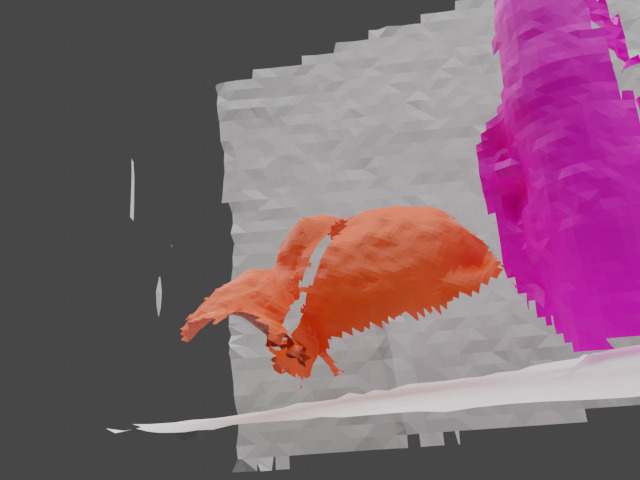} & \includegraphics[width=\linewidth]{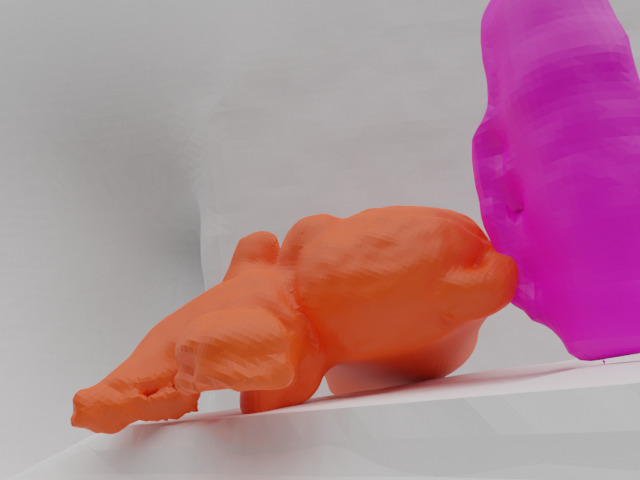} \\
		\includegraphics[width=\linewidth]{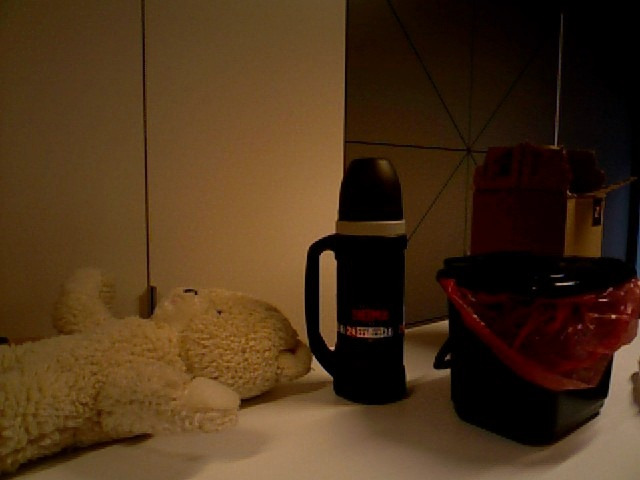}& \includegraphics[width=\linewidth]{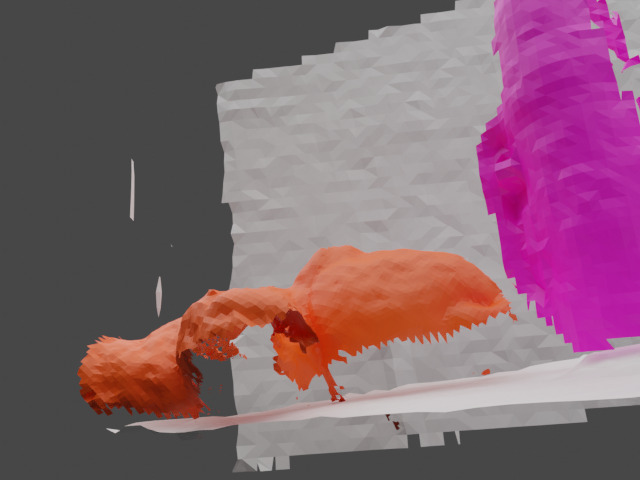} & \includegraphics[width=\linewidth]{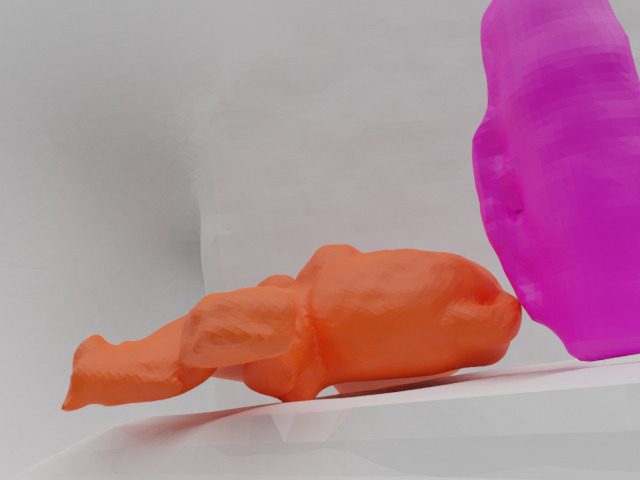}\\
		\vspace{.1ex} Input color & \vspace{.1ex} EM-Fusion & \vspace{.1ex} Ours\\
	\end{tabular}
	\caption{We propose a novel energy minimization approach to 3D reconstruction in dynamic scenes. Our approach uses a dynamic object-level SLAM method as frontend (such as EM-Fusion~\cite{strecke2019_emfusion}). Compared to the TSDF object maps of the frontend, our approach can leverage intersection constraints between objects and visibility constraints to complete object shapes in a physically plausible way. The top two rows show the same timestep in the scene \emph{car4} from different viewpoints. The bottom rows show different timesteps from the \emph{place-items} scene. See also the supplemental video for further visualization.}
\end{figure}

Humans have the remarkable ability to infer missing information using assumptions on the physical plausibility of scenes.
For instance, if we see objects lying on a table, we immediately conclude that the shape of the objects is constrained by the table surface.
We would be surprised if an object continues within the table.

In this paper we propose Co-Section, a novel approach to incorporate such constraints into a bottom-up 3D scene reconstruction method.
Our method tracks and maps dynamic objects in the scenes.
In dynamic scenes, however, we can use more than just the image measurements to reason about the object shapes: individual objects cannot penetrate each others while they move relative to each others and eventually are in contact.
In contrast to a variety of learning-based approaches for shape completion~\cite{firman2016_voxlets,song2017_ssc,dai2018_scancomplete,yang2019_recgan}, we incorporate such physical plausibility as intersection constraints in an energy-minimization framework.

Our formulation optimizes the implicit surfaces of the objects as a signed distance functions from oriented points.
The oriented point measurements are provided by a dynamic object-level SLAM frontend. It detects, segments, tracks and 3D reconstructs the objects in local volumetric SDF representations. 
Our energy minimization backend optimizes the object maps and incorporates intersection and hull constraints to complete the object shapes.

We demonstrate our approach on real and synthetic dynamic scene datasets.
We also assess our shape completion quantitatively using the ground-truth 3D models on the synthetic scenes.

In summary, our contributions are
\begin{itemize}
	\item We include intersection constraints for physically plausible shape completion in dynamic scenes.
This allows for inferring hidden surfaces of objects which can be useful for robotics or VR/AR applications.
To the best of our knowledge, we are the first to include such constraints for object-level 3D reconstruction in dynamic scenes.
	\item We demonstrate efficient optimization by initializing new object volumes coarse-to-fine and optimizing the volumes incrementally when new measurements are available.
	\item We assess our approach qualitatively and quantitatively on dynamic SLAM datasets. Our evaluation can serve as a baseline for future research in this area.
\end{itemize}

\section{Related work}
\subsection{Dense surface reconstruction}

Several approaches reconstruct 3D surfaces densely from point clouds~\cite{calakli2011ssd,kazhdan2006_poisson,Schroers_2014_variational,ummenhofer_2015_billionpoints}, 
depth images~\cite{curless96_volsdf} or RGB images from multiple view points~\cite{cremers2011_silh_stereo}. 
The methods typically consider the data as belonging to a single surface and do not reason about the surfaces of individual objects.

\textbf{Dense reconstruction from point clouds.}
Moving least squares surface reconstruction~\cite{Alexa_2003_crpss} fits a point set representing the surface to oriented point samples. 
Local parametrizations such as planar surface patches are assumed and the points are fit using moving least squares. 
Smooth signed distance surface reconstruction~\cite{calakli2011ssd} recovers a volumetric implicit surface representation from the oriented points using a variational energy minimization formulation.
The energy includes data terms favoring the points lying on the zero level set of the optimized implicit function and aligning its gradients with the surface normals.
A regularization term keeps the Hessian, \ie the second-order derivatives, of the implicit function small.
While we also use this regularization term, we use a data term that measures the distance of a voxel towards a point along the normal.
In Poisson surface reconstruction~\cite{kazhdan2006_poisson}, a smoothed indicator function is estimated whose gradient follows the surface normal field approximated with the oriented points.
The implicit function is recovered using a least squares approximation of the Poisson equation.
Implicit moving least squares formulations recover an implicit surface function at a set of points or voxel centers using a quadratic penalty on the deviation from local point samples along their normals.
Shen \etal \cite{Shen_2004_imls} estimate a fit of linear function coefficients to point samples.
Our formulation extends the Hessian-IMLS formulation of~\cite{Schroers_2014_variational} which combines a quadratic data penalty term with the L2-Hessian regularizer to estimate the signed distance function on a grid.
The approach by Ummenhofer \etal \cite{ummenhofer_2015_billionpoints} uses L1 norms on the data terms of SSD and a total variation regularizer on the gradient of the vector field which affords a primal-dual optimization scheme.
We favor the L2-Hessian regularization of the implicit function over total variation regularization of its vector field, since we are interested in  recovering smooth implicit surface functions without discontinuities in order to complete object shapes watertight.

\textbf{Dense reconstruction from depth and RGB images.}
Several approaches fuse depth images which can be recovered using multi-view stereo or active sensing principles.
A popular approach for incremental fusion of depth maps into volumetric signed distance functions is the seminal approach by Curless and Levoy~\cite{curless96_volsdf}.
Cremers and Kolev~\cite{cremers2011_silh_stereo} optimize the signed distance function directly from intensity images using multi-view stereo and silhouette constraints.
These dense reconstruction methods are restricted to static scenes and do not incorporate intersection constraints between objects like our method.

\subsection{Dense object-level 3D reconstruction in dynamic environments}
 
Several methods have been proposed that track and map moving objects with RGB-D cameras~\cite{ruenz_2017_co-fusion,ruenz_2018_maskfusion,xu_2019_midfusion,Hachiuma2019_detectfusion,strecke2019_emfusion}.
None of the methods, however, considers mutual constraints between the objects for inferring hidden surfaces at their intersections.
Some recent works also consider interactions between objects, for instance, between human bodies and the static environment~\cite{Hassan_2019_sceneconstraints}, or hands and objects~\cite{hasson2019_handsobjects}.
However, these methods require learned parametric deformation models of bodies and hands.
They furthermore rely on deep learning for pose and shape regression from images.

\subsection{Shape completion}
Various approaches have been proposed that learn to complete volumetric reconstructions obtained with depth cameras~\cite{firman2016_voxlets,song2017_ssc,dai2018_scancomplete,yang2019_recgan}.
Firman \etal \cite{firman2016_voxlets} train random forests to map depth image patches to local 3D voxel representations.
Song \etal \cite{song2017_ssc} complete voxel representation of single depth images.
They train a 3D convolutional neural network end-to-end on a synthetic 3D scene dataset.
ScanComplete~\cite{dai2018_scancomplete} applies 3D CNNs in a hierarchical fashion to complete and semantically label voxel grids of point clouds fusing multiple views.
Yang \etal \cite{yang2019_recgan} propose a GAN-style approach to predict completed shapes from depth images.
The recently proposed X-Section~\cite{nicastro_2019_xsection} predicts the end-point of an object along a ray which can be used with volumetric SDF fusion~\cite{curless96_volsdf} to obtain completed shapes. 

Our energy minimization approach includes physically plausible constraints which do not need to be learned from data.
It thus provides an orthogonal approach that might lead to promising combinations with these learning-based approaches in future research.

\section{Method}
Our goal is to use depth measurements that are associated to object-level 3D maps in order to generate a globally consistent 3D model for each object.
Our method reasons about occluded 3D surfaces by integrating intersection constraints between the objects.
The key idea behind this is that dynamic objects reveal some information about their maximum extend in the viewing direction if they pass in front of another known surface.
We use this information to infer where an objects ends in directions that have not been directly observed.

Our approach uses a frontend which performs dynamic object-level RGB-D SLAM.
The frontend provides a segmentation of the scene into dynamic objects.
It estimates the object trajectories and 3D maps which are fused from the depth measurements on the objects.
Our main novelty in this work is an energy minimization backend which consolidates the dense 3D reconstructions of the moving objects with physical plausibility constraints between the objects.
To this end, it takes intersection constraints between the objects into account.
We also include hull constraints which limit the object surface with the observed free space.

\subsection{Frontend}
\begin{figure*}
	\centering
	\includestandalone[width=.99\linewidth,build={latexoptions={-output-directory=./figures -recorder}}]{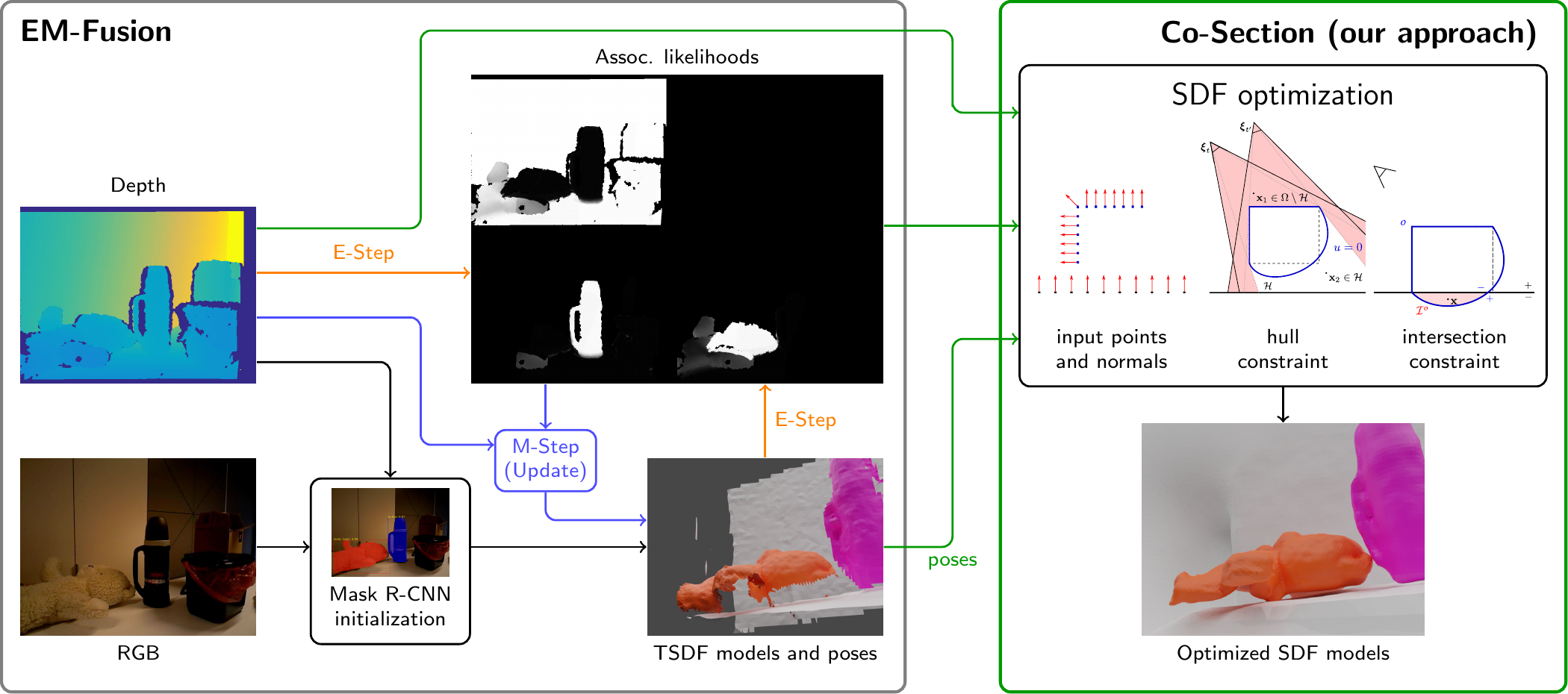}
	\caption{Our approach uses input depth maps and the estimated object and camera poses as well as the pixel-wise object association likelihoods from EM-Fusion \cite{strecke2019_emfusion} in a global optimization framework to retrieve optimized SDF models that take physical plausibility constraints into account. EM-Fusion uses Mask R-CNN \cite{he_2017_maskrcnn} segmentations to initialize objects and subsequently tracks and maps them in an EM-like framework by estimating geometric association likelihoods from existing models and input depth maps.}
	\label{fig:overview}
\end{figure*}

The dynamic SLAM frontend associates depth measurements to object-level maps.
RGB-D SLAM methods can be used for this purpose that track and reconstruct moving objects such as \cite{ruenz_2017_co-fusion,ruenz_2018_maskfusion,xu_2019_midfusion,Hachiuma2019_detectfusion,strecke2019_emfusion}.
We choose to build upon our dynamic SLAM approach EM-Fusion \cite{strecke2019_emfusion} and extend it to generate globally consistent object models, including priors about object intersections.
In EM-Fusion, we represent individual objects as truncated signed distance fields (TSDFs).
We find the signed distance field (SDF) representation especially useful for the goal of including intersection constraints since it contains distance information to the closest surfaces.
Additionally, it gives information whether a 3D point is inside or outside an object by negative or positive values, respectively.

In EM-Fusion, objects are detected by Mask R-CNN \cite{he_2017_maskrcnn} and initialized as per-object TSDF volumes.
For correct associations of depth pixels to object models, in each frame a geometric association likelihood $a^o$ is estimated and used as a per-pixel weight for tracking and mapping the different models using the input depth maps.
New depth measurements in EM-Fusion are integrated by weighted average fusion \cite{curless96_volsdf} with a cap on the maximum weight as in \cite{newcombe2011_kinectfusion}.
This approach has been proven to yield a maximum likelihood estimate of the surface \cite{curless96_volsdf}.
However, this method cannot map thin structures and edges of objects well if they are viewed from different sides.
Previously observed surfaces might be overridden with later measurements from the other side of the object, if they lie within the truncation distance.
To alleviate these problems, we store depth frames as keyframes in equidistant timesteps and compute oriented pointclouds (consisting of points $\mathbf{p}_i$ and their corresponding normals $\mathbf{n}_i$) from them, given the poses estimated by EM-Fusion.
Then we perform a global optimization of the signed distance field from this pointcloud to generate a globally consistent model.
An overview of the structure of EM-Fusion and what data we use in our optimization is given in Fig.~\ref{fig:overview}.
We will now explain the details of the optimization framework we employ.

\subsection{Global SDF optimization}
\textbf{Energy formulation.}
Several approaches have been proposed for global optimization of signed distance fields (SDFs) from oriented pointclouds, \eg, \cite{calakli2011ssd,kazhdan2006_poisson}.
Schroers \etal \cite{Schroers_2014_variational} proposed a taxonomy and generalization of these approaches.
In our work, we adopt the Hessian-IMLS energy formulation by Schroers \etal \cite{Schroers_2014_variational}.
The overall energy that we optimize is the sum of four parts which we will explain in the following:
\begin{equation}
	E(u) = E_\mathrm{data}(u) + E_\mathrm{reg}(u) + E_\mathrm{hull}(u) + E_\mathrm{inter}(u).
	\label{eq:energy}
\end{equation}

Given measurement points $\mathbf{p}_i$ and their corresponding normals $\mathbf{n}_i$, Hessian-IMLS defines the energy $E$ over possible SDFs $u : \Omega \to \mathbb{R}$ as
\begin{equation}
	\begin{split}
	E_\mathrm{Hessian-IMLS}(u) &= E_\mathrm{data}(u) + E_\mathrm{reg}(u)\\
	E_\mathrm{data}(u)&= \int_{\Omega}\sum_{i=1}^{N} w_i ( u(\mathbf{x}) - f_i(\mathbf{x}) )^2 ~\mathrm{d}\mathbf{x}\\
	E_\mathrm{reg}(u)&= \alpha \int_{\Omega} \left\|\mathbf{H}u(\mathbf{x})\right\|_F^2 ~\mathrm{d}\mathbf{x}
	\end{split}
\label{eq:data_energy}
\end{equation}
with $\Omega \subset \mathbb{R}^3$.
Here, $f_i(\mathbf{x}) = \langle \mathbf{x} - \mathbf{p}_i, \mathbf{n}_i \rangle$ denotes the signed distance measurement for point $\mathbf{p}_i$ along the surface normal $\mathbf{n}_i$.
The weight $w_i(\mathbf{x}) = w_\sigma(\|\mathbf{x} - \mathbf{p}_i\|) \cdot a^o(\mathbf{p}_i)$ is computed as the product of a Gaussian weight $w_\sigma(s) = \exp \left( (s/\sigma) ^2\right)$ reducing the influence of points for voxels further from the surface as in \cite{Schroers_2014_variational} and the association likelihood $a^o(\mathbf{p}_i)$ of point $\mathbf{p}_i$ to object $o$ from EM-Fusion.
We set $\sigma$ equal to the voxel size in our experiments and cap the weights when the distance to the measurement is larger than $3\sigma$.
The L2-regularizer on the Hessian of $u$, $\|\mathbf{H}u\|_F^2$ follows the intuition that the distance to the closest surface changes approximately linearly with a change in position.

\begin{figure}
	\begin{center}
		\centering
		\includegraphics[width=0.99\linewidth]{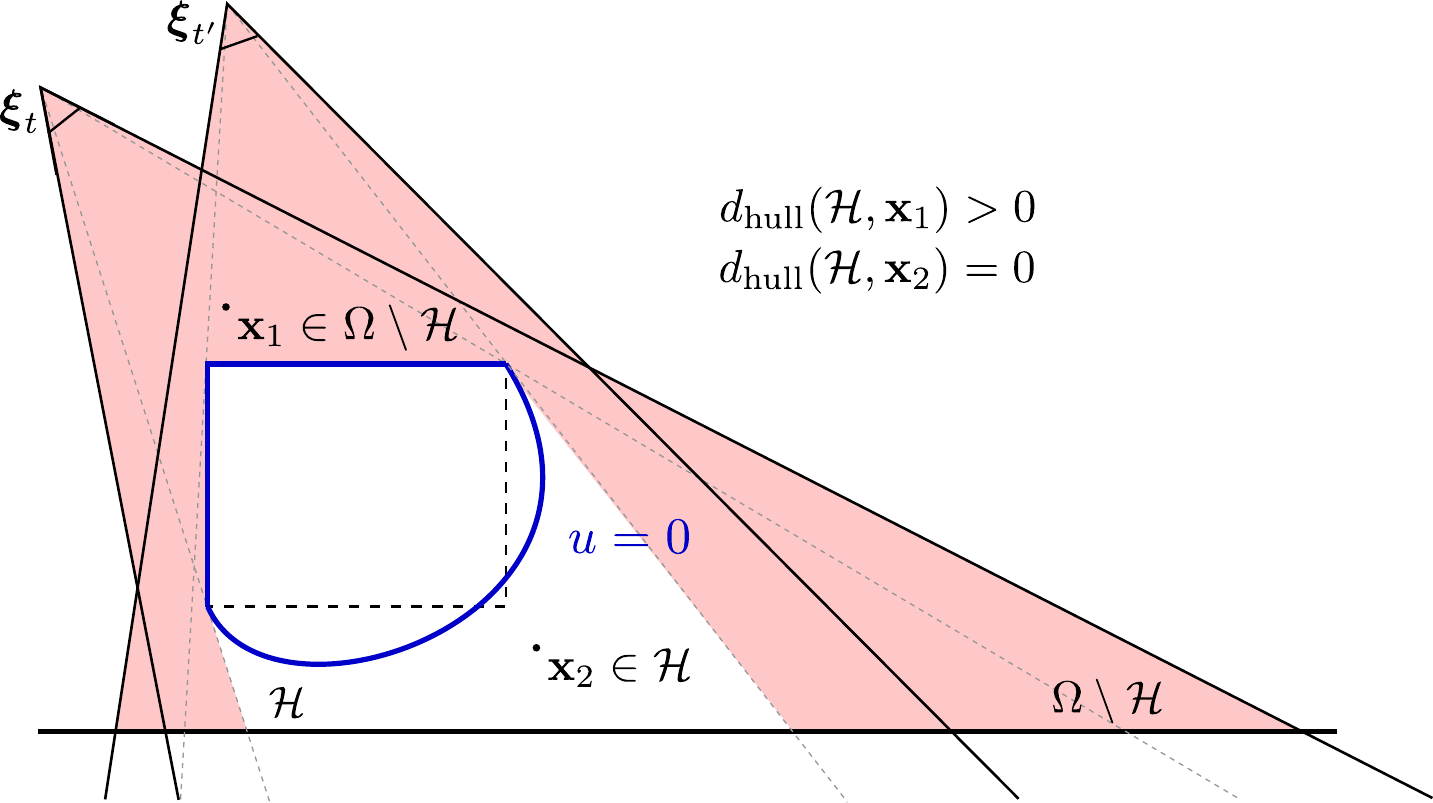}
	\end{center}
	\caption{The hull constraint effectively limits the reconstructed surface within the observed free-space.}
	\label{fig:hullconstraint}
\end{figure}

\textbf{Hull constraint.}
Schroers \etal also propose a hull constraint which can be included into Hessian-IMLS,
\begin{equation}
	E_\mathrm{hull}(u) = \beta_\mathrm{hull} \int_{\Omega\setminus \mathcal{H}} \max\{ 0, d_\mathrm{hull}(\mathcal{H},\mathbf{x}) - u(\mathbf{x}) \}^2 ~\mathrm{d}\mathbf{x},
	\label{eq:hullconstr}
\end{equation}
where $d_\text{hull}(\mathcal{H}, \mathbf{x})$ denotes the distance of point $\mathbf{x}$ to the hull $\mathcal{H}\subset \mathbb{R}^3$ in which the reconstruction lies.
Essentially, this constrains the reconstructed implicit surface to a reasonable area and proves beneficial in their experiments \cite{Schroers_2014_variational}.
Since the data we collect is not just the raw oriented point cloud, but depth measurements over time, we implement the hull $\mathcal{H}$ to consist of all unseen parts of the scene (hidden behind seen surfaces or outside the field of view).
Furthermore, to avoid the potentially expensive exact computation of $d_\mathrm{hull}(\mathcal{H},\mathbf{x})$, we use the voxel size as a lower bound for all points $\mathbf{x}\notin \mathcal{H}$.
This effectively limits the reconstructed surface with the observed free space.
See Fig.~\ref{fig:hullconstraint} for an illustration of the hull constraint. 

\textbf{Intersection constraint.}
We propose to implement an intersection constraint for multiple objects with a similar form like the hull constraint in equation \eqref{eq:hullconstr}.
For each object $o$ in the set of objects $\mathcal{O}$, we define the intersection distance $d_\mathrm{inter}$ based on the observations of multiple (moving) objects in several timesteps $t$:
\begin{equation}
	d_\mathrm{inter}^o(\mathbf{x}) = 
	\max \left\{ \max_{t\in T, p\in\mathcal{O}}\left\{ -u^p(\mathbf{x}_t) \right\}, 0 \right\}.
	\label{eq:d_inter}
\end{equation}
Here, $\mathbf{x}_t = \mathbf{T}(\boldsymbol{\xi}^p_t)^{-1}\mathbf{T}(\boldsymbol{\xi}^o_t) \, \mathbf{x}$ denotes the point $\mathbf{x}$ transformed from the coordinate system of object $o$ to object $p$ at timestep $t$ using the poses of $o$ and $p$ at that timestep, $\boldsymbol{\xi}_t^o$ and $\boldsymbol{\xi}_t^p$, respectively.
Intuitively, this distance measures the maximum penetration of a point $\mathbf{x}$ of object $o$ in object $p$ over all time steps. 

\begin{figure*}
	\begin{center}
		\centering
		\includegraphics[width=0.91\textwidth]{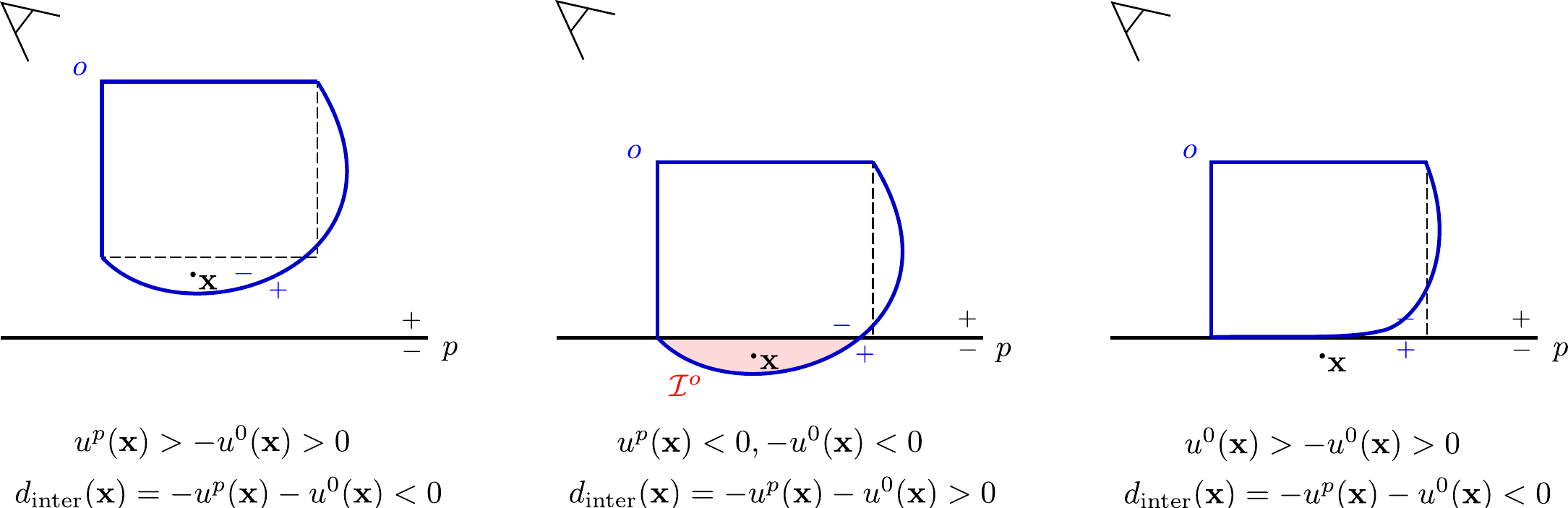}
	\end{center}
	\caption{Our intersection constraint is based on a measure of interpenetration distance~$d_{\text{inter}}$. It penalizes signed distance functions that interpenetrate each other. Left: interpenetration distance for two objects that do not intersect each others. Middle: case of two intersecting objects. Right: Distance after resolved interpenetration.}
	\label{fig:intersectionconstraint}
\end{figure*}

Using this distance, we define the energy term
\begin{equation}
	E_\mathrm{inter}(u^o) = \beta_\mathrm{inter} \int_{\mathcal{I}^o} \max\{ 0, d_\mathrm{inter}^o - u^o \}^2 ~\mathrm{d}\mathbf{x},
	\label{eq:interconstr}
\end{equation}
where $\mathcal{I}^o = \left\{ \mathbf{x} \in \Omega \mid d_\mathrm{inter}^o(\mathbf{x}) > 0 \right\}$.
This term favors SDFs that do not interpenetrate each others.
Fig.~\ref{fig:intersectionconstraint} illustrates the intersection constraint in several situations. 

\textbf{Optimization.}
Since all parts of the energy are convex and differentiable in $u$, we follow Schroers \etal \cite{Schroers_2014_variational} and implement the optimization with the Fast Jacobi algorithm proposed by Weickert \etal \cite{Weickert_2015_cyclic}.

Since in our case the data measurements are sparse compared to the overall volume size, optimization of the background volume on the finest resolution of $256^3$ only converges slowly for the first run.
We thus propose to initialize each volume with a coarse-to-fine optimization by optimizing only $E_\mathrm{data}$ from equation \eqref{eq:data_energy} first on a resolution of $32^3$ and subsequently upsampling the result by splitting each voxel into 8 and using this as initialization for the next level.
Even without interpolation, this strategy yielded reasonable fast and accurate results.
For the finest level, we optimize equation \eqref{eq:energy} and include the hull and intersection constraints.
The coarse-to-fine scheme is only applied for the first optimization of each volume.
In our experiments, the previous optimization result can adapt reasonably fast to new measurements in later optimization runs.
Furthermore, we found the application of a coarse-to-fine scheme in later time steps to actually slow down the optimization in some cases.
This is caused by removing fine details at lower resolutions which were already recovered in the previous optimization run.

\textbf{Wrong data and inaccurate detections.}
The detection of dynamic objects in \cite{strecke2019_emfusion} relies on Mask R-CNN \cite{he_2017_maskrcnn}.
The detection is carried out at low framerate.
This might lead to dynamic objects being missed when they move into view.
Consequently, point associations with the background or other object models are incorporated which are difficult to remove later.
Since EM-Fusion is mainly designed to track rigid objects, it is fair to assume that space that was seen unoccupied before in a model will not be occupied by the same object in the future.
We thus do not allow association of points in this space with the respective model and set $f_i(\mathbf{x}) = w_i(\mathbf{x}) = 0$ for $\mathbf{x}\in\Omega\setminus\mathcal{H}$.
Note that this way, the hull $\mathcal{H}$ will only get smaller over time.

Furthermore, EM-Fusion yields balanced association likelihoods for the background model and objects that remain static since the first view.
Including these points in both the object and the background model during our optimization would yield conflicting dataterms for the intersection constraint.
To avoid this, we use the volumetric foreground probability estimated by EM-Fusion and remove all point measurements in the foreground region of an object model from all other models.
While this strategy helps with handling static objects, it might cause dents when models touch each others since it might also remove correct point measurements from the other model.
This strategy might not be needed in scenes where no static objects are detected by EM-Fusion.

\textbf{Ordering the optimization of different volumes.}
The intersection constraint as formulated in equation \eqref{eq:d_inter} introduces a chicken-and-egg-problem.
The optimization for object $o$ depends on the optimization results for all other models while those results depend on the optimization result of $o$.
To resolve this, we observe that planar surfaces (as they are common for \eg walls and floors in manmade environments) can be completed reasonably well using the Hessian norm prior in equation \eqref{eq:data_energy}.
Thus to compute the intersection constraint in Eq.~\eqref{eq:d_inter}, we use the optimized SDF of the background model and approximate $u^p$ in the case of object models using the point measurements, i.e. $u^p(\mathbf{x}) \approx \frac{1}{N} \sum_{i=1}^N w_i(\mathbf{x})f_i(\mathbf{x})$.

\section{Experiments}
We evaluate our approach in qualitative and quantitative experiments on the dataset provided with Co-Fusion~\cite{ruenz_2017_co-fusion}.
We use 2 real and 1 synthetic scene in which up to 3 objects move independently.
Besides qualitative results, we also provide quantitative results on shape completion which we obtain in the synthetic scene using the ground-truth meshes of the objects.
In an ablation study, we analyze the effectiveness of the individual components of our energy minimization approach.

We chose the parameters for weighting the different parts of the energy empirically as $\alpha=0.005$, $\beta_\mathrm{hull} = \beta_\mathrm{inter} = 0.001$, and choose a cycle length of $20$ for the Fast Jacobi optimizer \cite{Weickert_2015_cyclic}.
We used a single set of parameters throughout our experiments.

\subsection{Qualitative Results}

\begin{figure*}
\footnotesize
	\begin{center}
	\setlength{\tabcolsep}{1pt}
	\renewcommand{\arraystretch}{0.6}
	\begin{tabular}{P{.158\linewidth}P{.158\linewidth}P{.158\linewidth}P{.158\linewidth}P{.158\linewidth}P{.158\linewidth}}
	  Input color\vspace{1ex} & TSDF & baseline & baseline + hull & baseline + intersection & Co-Section (ours)\\
		\includegraphics[width=\linewidth]{figures/car4/Color0142} & \includegraphics[width=\linewidth]{figures/car4/wide/truck2_tsdf/0142} & \includegraphics[width=\linewidth]{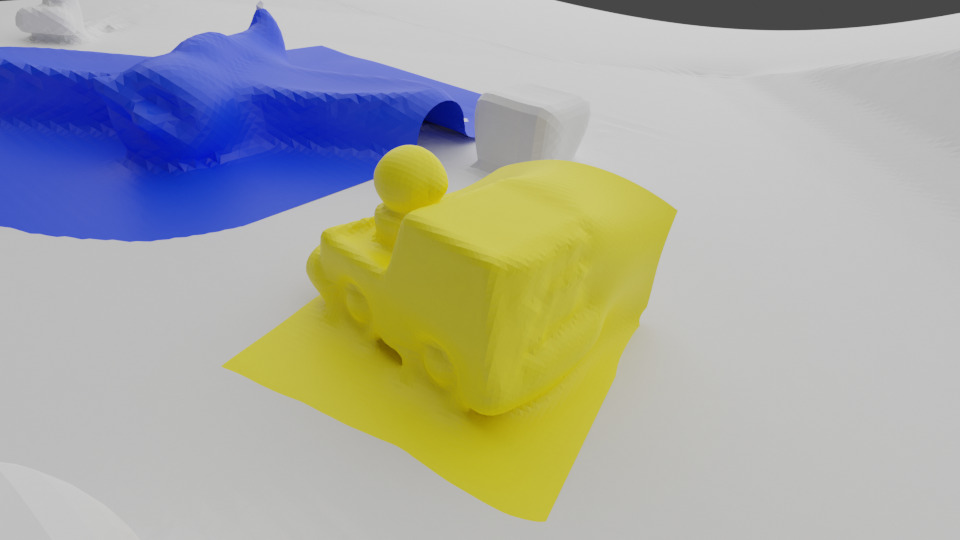} & \includegraphics[width=\linewidth]{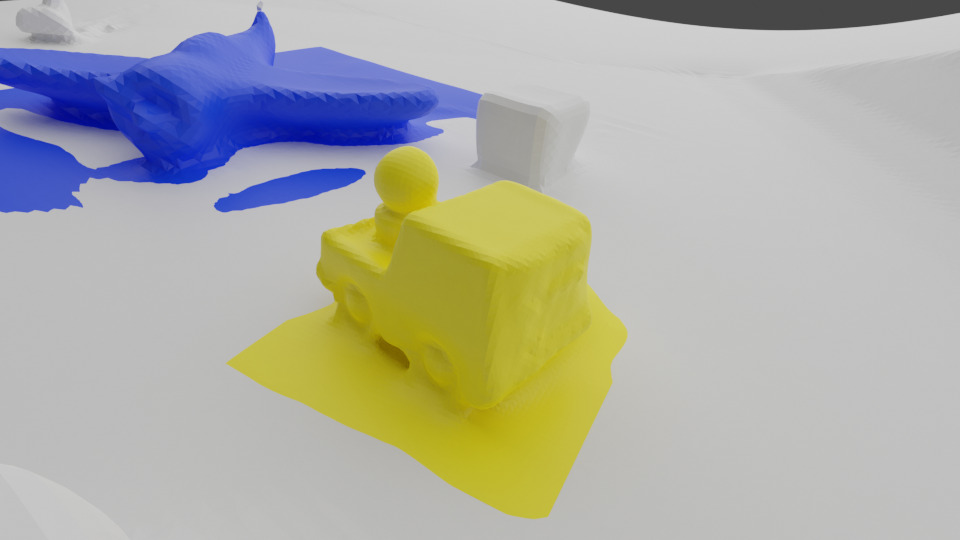} & \includegraphics[width=\linewidth]{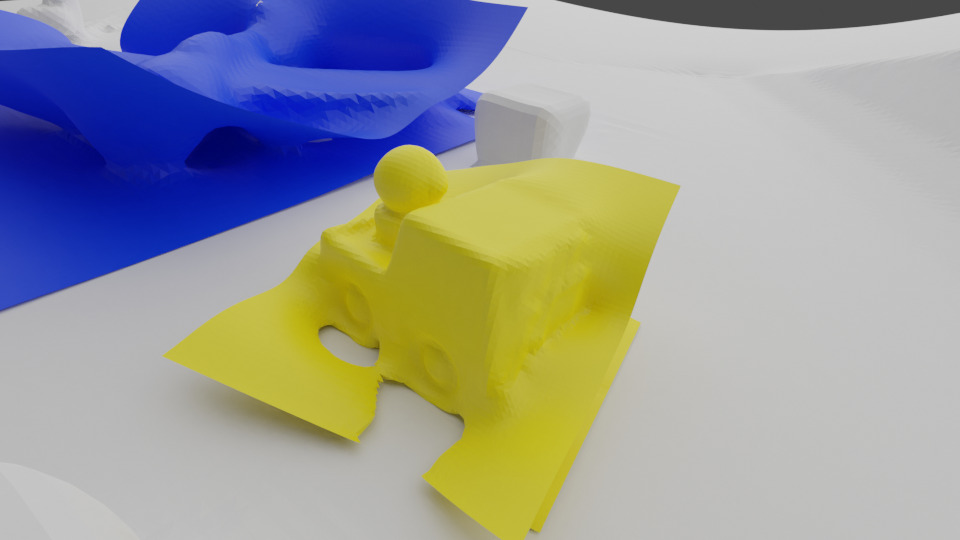} & \includegraphics[width=\linewidth]{figures/car4/wide/truck2_final/0142}\\
		& \includegraphics[width=\linewidth]{figures/car4/wide/truck1_tsdf/0142} & \includegraphics[width=\linewidth]{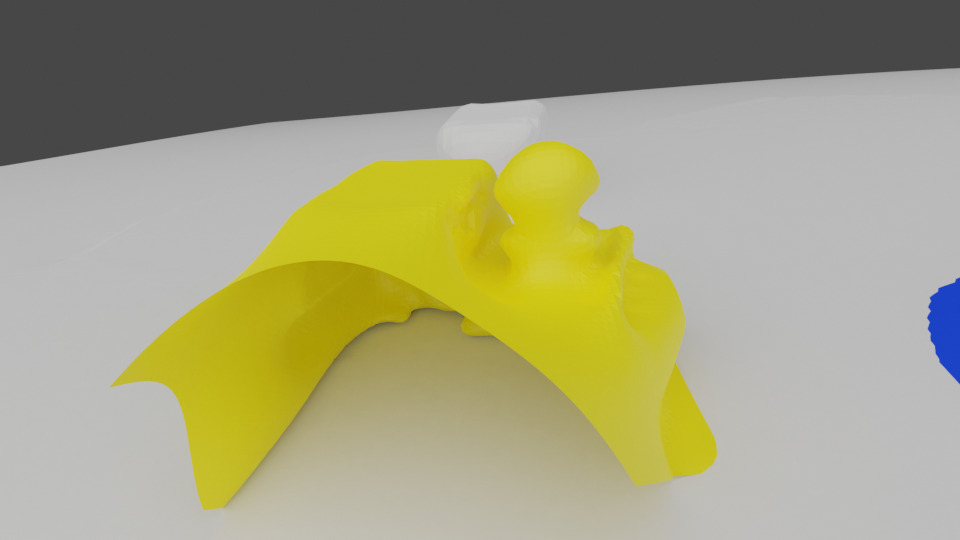} & \includegraphics[width=\linewidth]{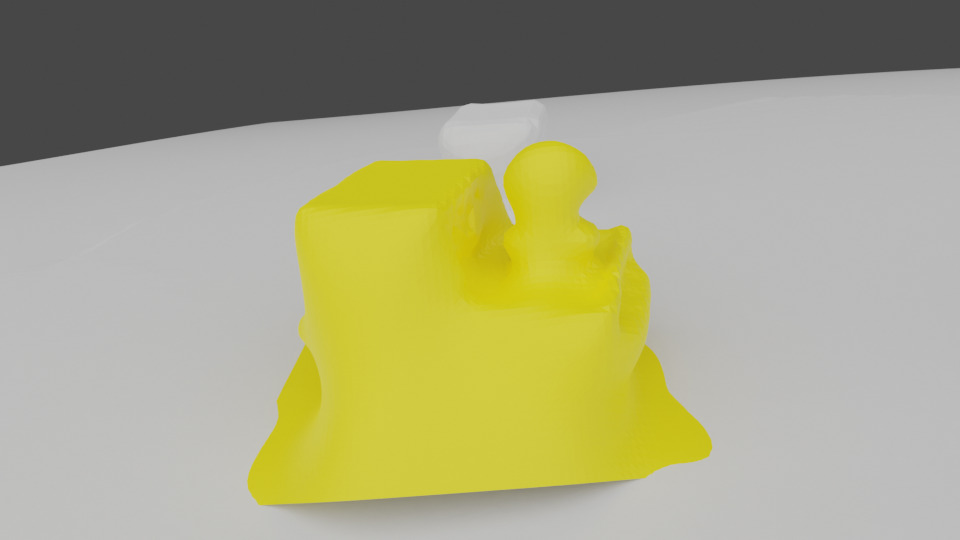} & \includegraphics[width=\linewidth]{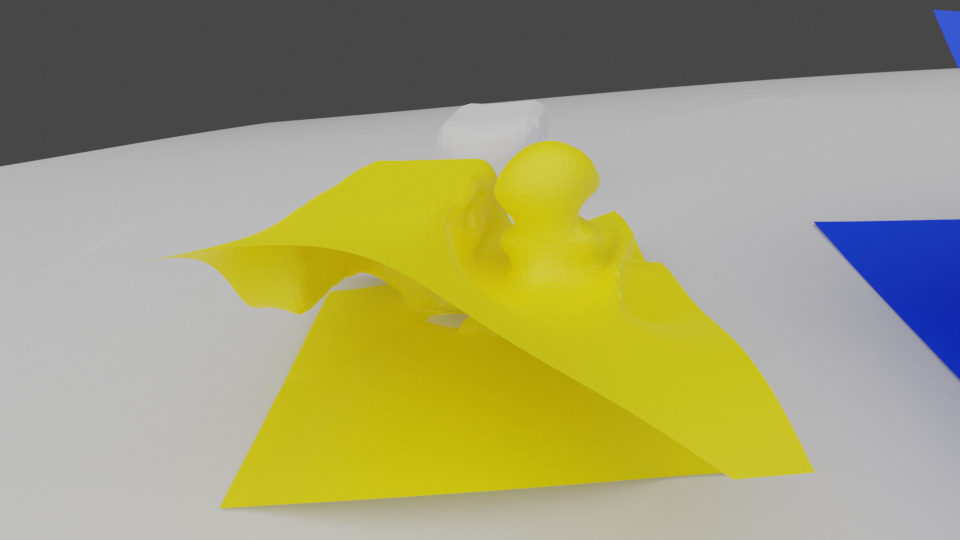} & \includegraphics[width=\linewidth]{figures/car4/wide/truck1_final/0142}\\
		\includegraphics[width=\linewidth]{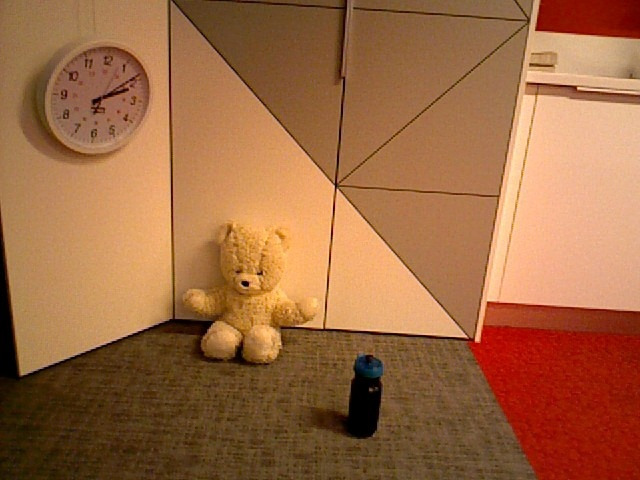} & \includegraphics[width=\linewidth]{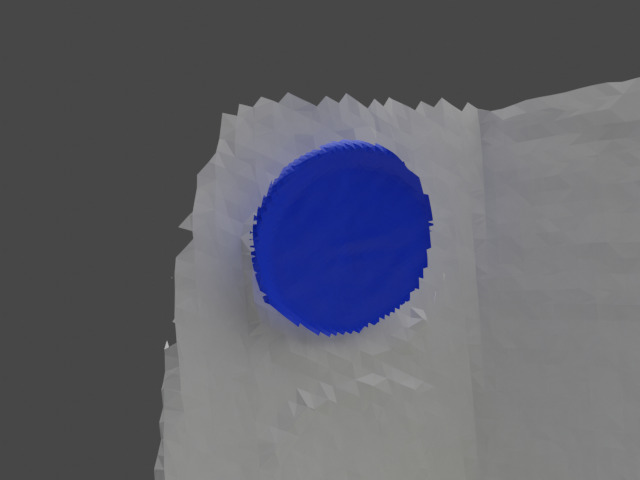} & \includegraphics[width=\linewidth]{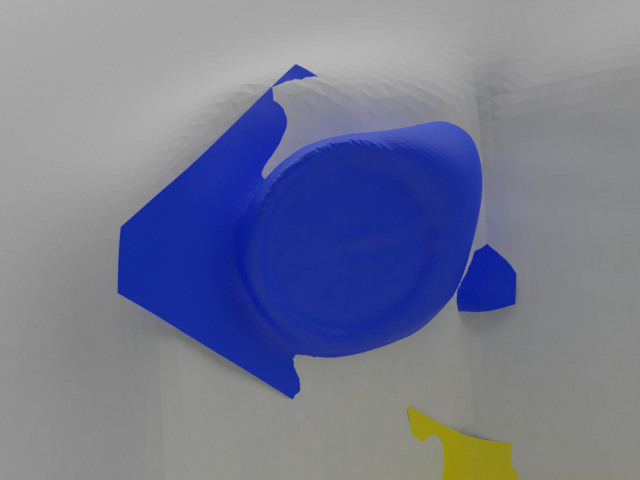} & \includegraphics[width=\linewidth]{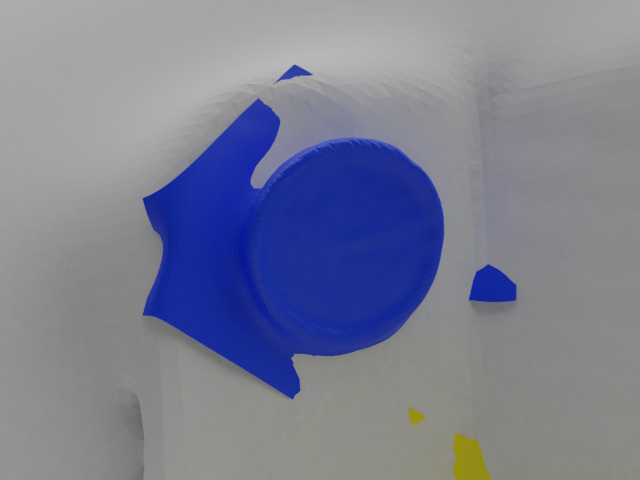} & \includegraphics[width=\linewidth]{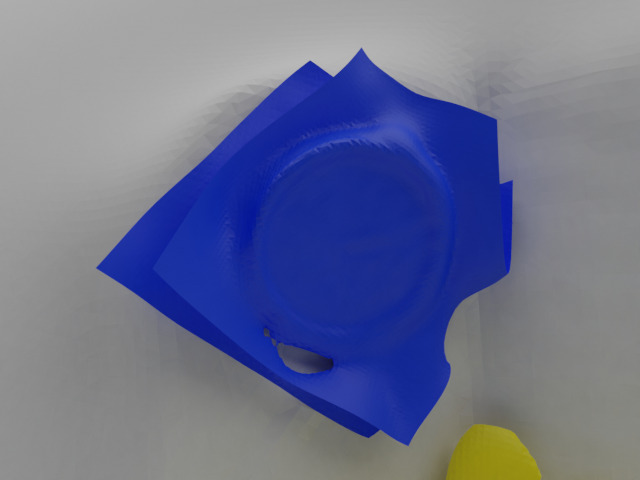} & \includegraphics[width=\linewidth]{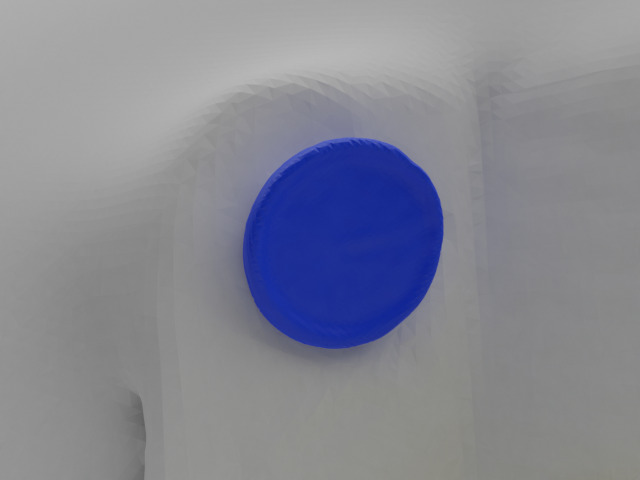}\\
		& \includegraphics[width=\linewidth]{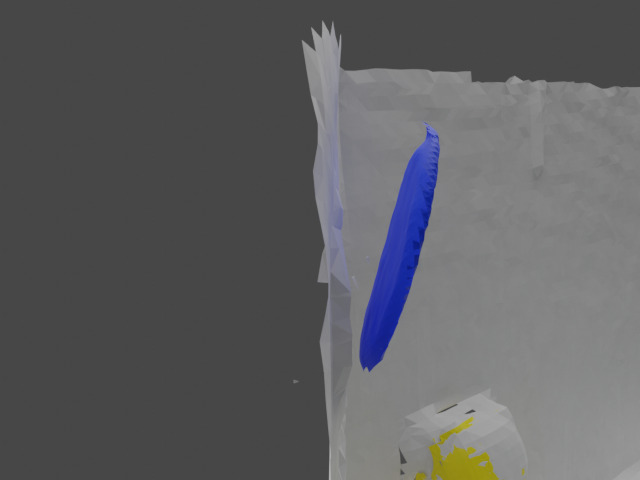} & \includegraphics[width=\linewidth]{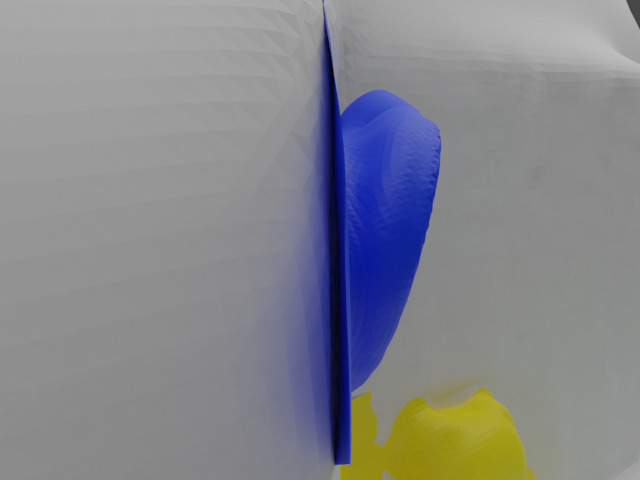} & \includegraphics[width=\linewidth]{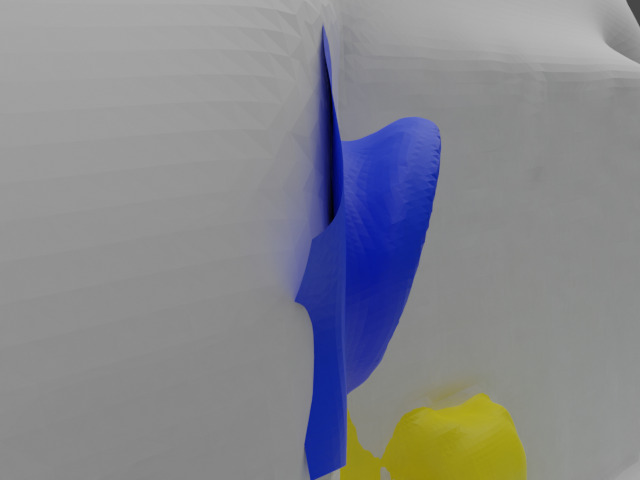} & \includegraphics[width=\linewidth]{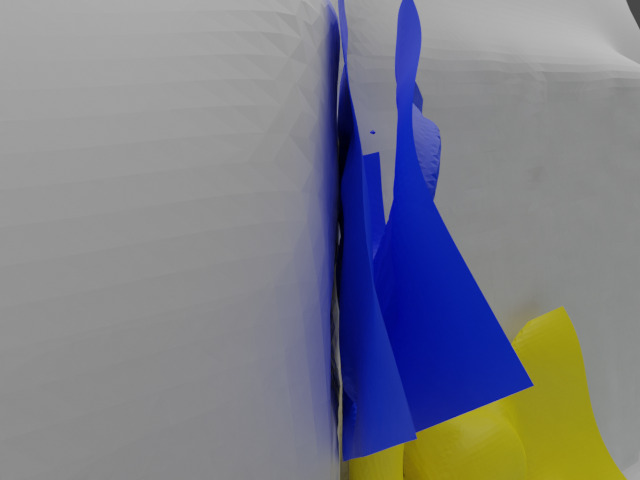} & \includegraphics[width=\linewidth]{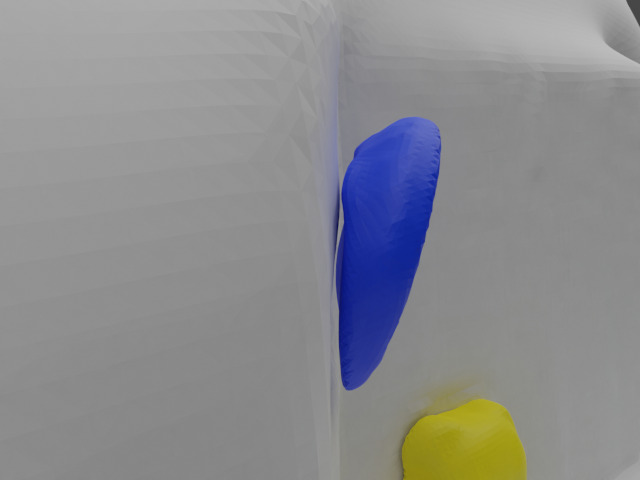}\\
		\includegraphics[width=\linewidth]{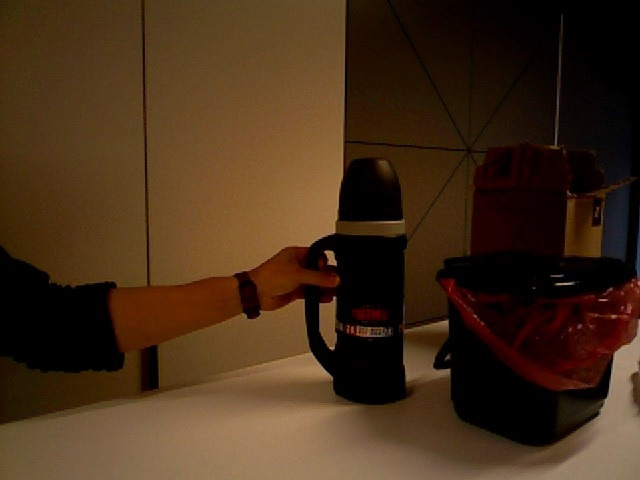} & \includegraphics[width=\linewidth]{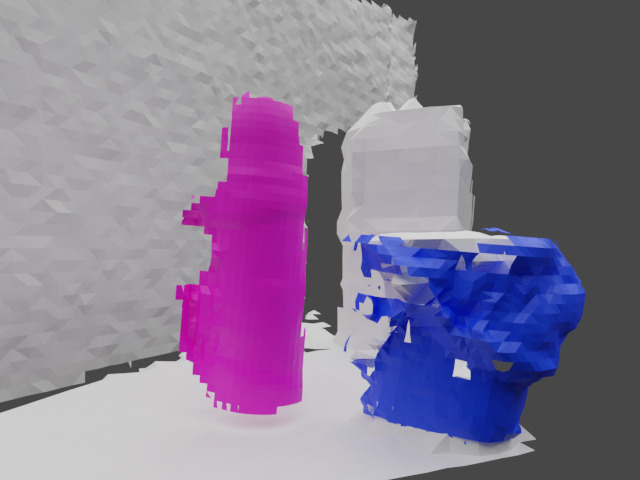} & \includegraphics[width=\linewidth]{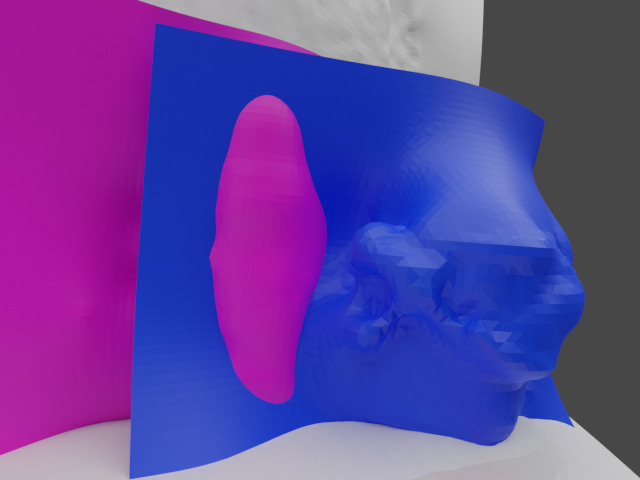} & \includegraphics[width=\linewidth]{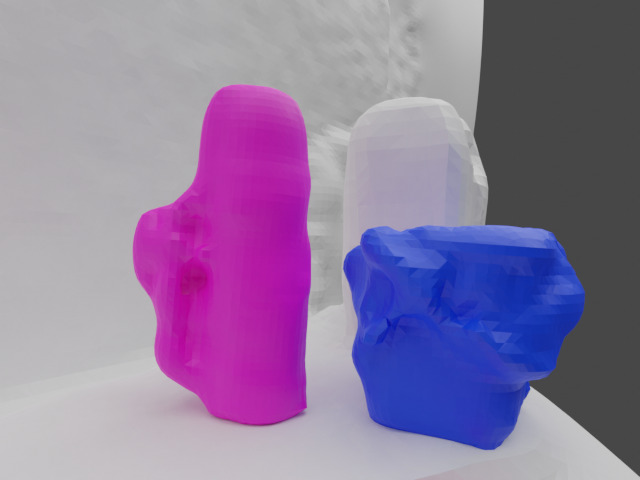} & \includegraphics[width=\linewidth]{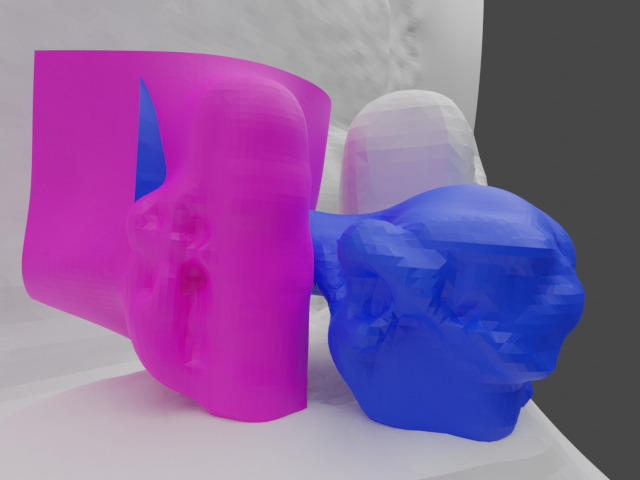} & \includegraphics[width=\linewidth]{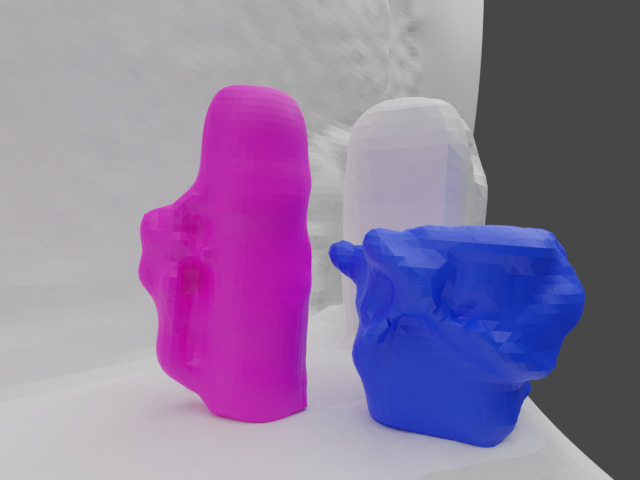}\\
		& \includegraphics[width=\linewidth]{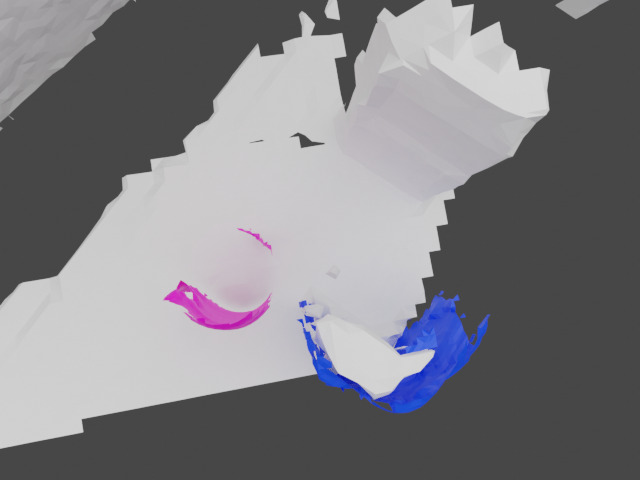} & \includegraphics[width=\linewidth]{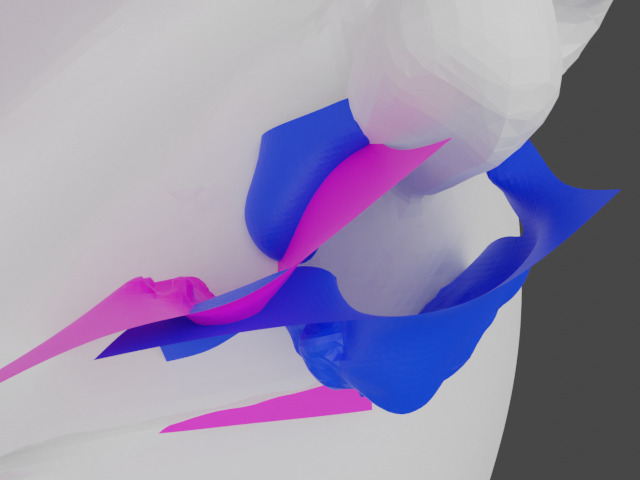} & \includegraphics[width=\linewidth]{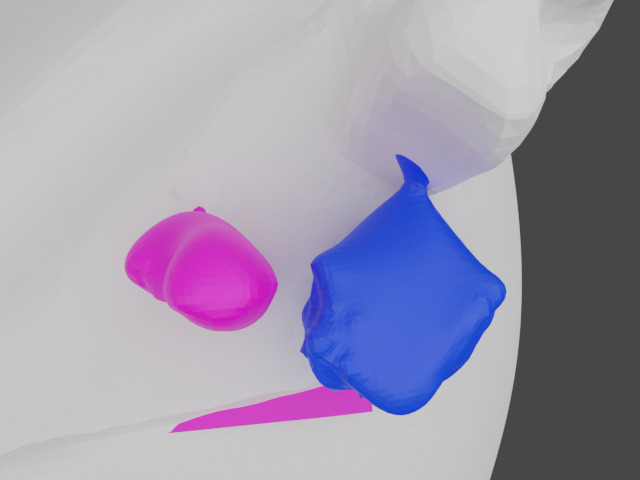} & \includegraphics[width=\linewidth]{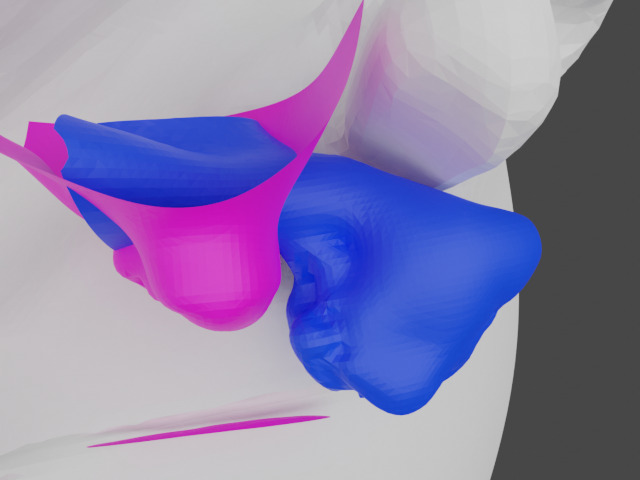} & \includegraphics[width=\linewidth]{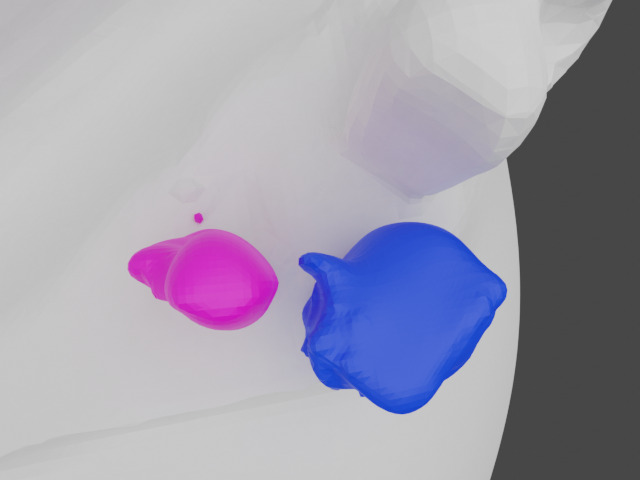}\\
		\includegraphics[width=\linewidth]{figures/place-items/Color0859} & \includegraphics[width=\linewidth]{figures/place-items/teddy_tsdf/0859} & \includegraphics[width=\linewidth]{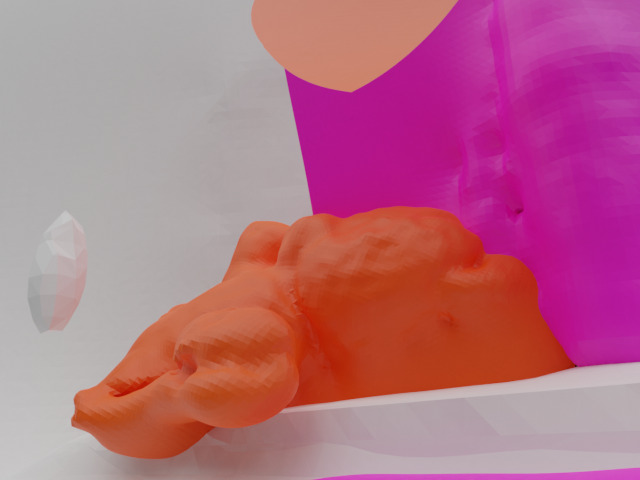} & \includegraphics[width=\linewidth]{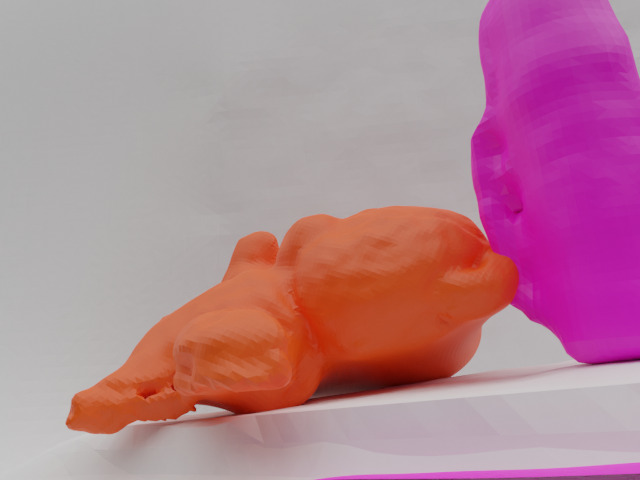} & \includegraphics[width=\linewidth]{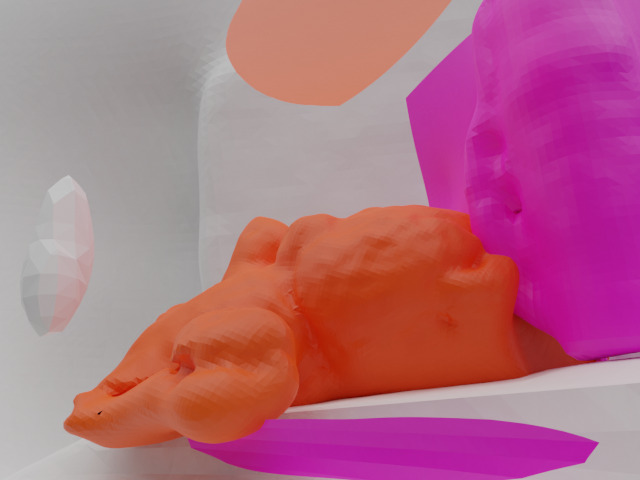} & \includegraphics[width=\linewidth]{figures/place-items/teddy_final/0859}\\
		\includegraphics[width=\linewidth]{figures/place-items/Color0870} & \includegraphics[width=\linewidth]{figures/place-items/teddy_tsdf/0870} & \includegraphics[width=\linewidth]{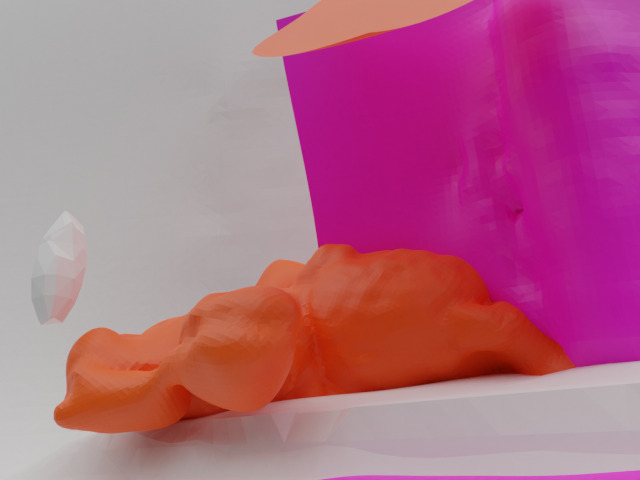} & \includegraphics[width=\linewidth]{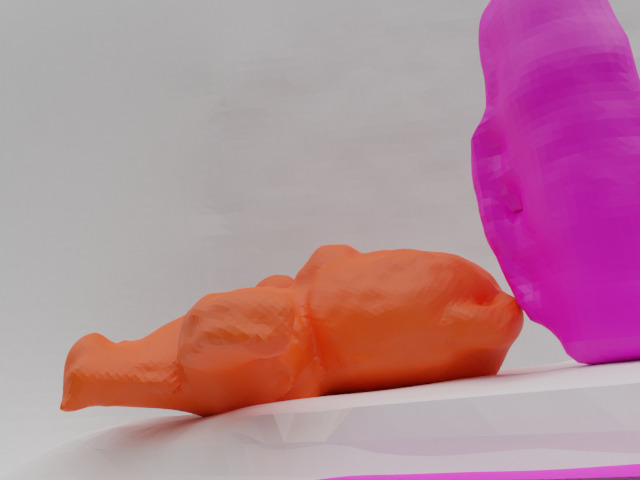} & \includegraphics[width=\linewidth]{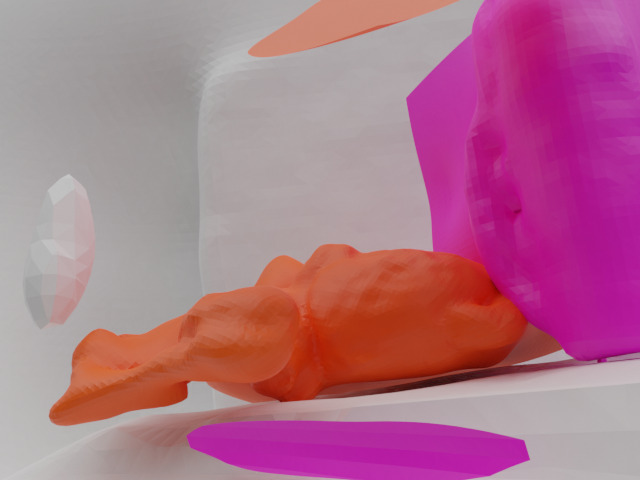} & \includegraphics[width=\linewidth]{figures/place-items/teddy_final/0870}\\
	\end{tabular}
	\end{center}
	\caption{Qualitative object shape reconstruction results on Co-Fusion sequences. From top to bottom: truck (car4), clock (sliding-clock), bottle and dustbin (place-items), and teddy (place-items). The TSDF shape only includes the observed part of the objects. While the hull constraint limits the surface in the observed free space, the intersection constraint closes the object shape at intersections with other shapes. Our full approach recovers closed surfaces which approximate the actual object shapes. Note that we do not show ``person'' objects such as the arm in the bottle and dustbin example.}
	\label{fig:qualrecon}
\end{figure*}

We provide qualitative results of our approach in Fig.~\ref{fig:qualrecon}.
The first three examples show different viewpoints of the same time step, while the last one (placement of teddy) illustrates how the shape changes over time. 
The TSDF reconstructions estimated by EM-Fusion can only take the visible part of the object into account.
When using our global optimization approach without hull and intersection constraints, the object surface is continued as smooth as possible towards the borders of the object map.
The hull constraints wraps the surfaces such that they are limited by the observed free space.
However, the surface can freely intersect with the background surface under the object.
When including the intersection constraint, this penetration is avoided.
However, without the hull constraint, the gradient introduced by the intersection constraint might be continued in regions outside the actual objects, causing unwanted zero-crossings (see column 4 in Fig.~\ref{fig:qualrecon}).
In combination with the hull constraint, these unwanted zero-crossings are removed by penalizing negative SDF values in observed free space.
Thus, the object shape is closed and approximates the actual object shape well.

More qualitative examples including the development of the optimized surfaces over time can be found in the supplemental video.

\subsection{Quantitative Results}
\textbf{Evaluation metric.}
We evaluate shape reconstruction using the measures and tools suggested in~\cite{Stutz2018CVPR}\footnote{https://github.com/davidstutz/mesh-evaluation}.
The method samples 10{,}000 points on the mesh reconstructed from the SDFs and the ground-truth mesh uniformly.
It then computes the average distance of each sample point in the reconstructed mesh to the ground-truth mesh (point-to-triangle distance) as \emph{accuracy}, and the average distance of each sample from the ground-truth mesh to the reconstructed mesh as \emph{completeness}.
Unfortunately, for this evaluation the alignment of the ground-truth mesh is not provided by the dataset.
Thus, we manually align scale and pose of the ground-truth meshes with our output meshes by selecting point correspondences on the meshes.
Note that, we only need to align each object once for each optimized map (using the full approach), since all reconstructions are based on the EM-Fusion result and share the same pose estimate.

\begin{figure*}
\footnotesize
	\begin{center}
	\setlength{\tabcolsep}{0pt}
	\renewcommand{\arraystretch}{0.6}
	\begin{tabular}{P{.189\linewidth}P{.189\linewidth}P{.189\linewidth}P{.189\linewidth}P{.189\linewidth}P{0.055\linewidth}}
	  TSDF & baseline & baseline + hull & baseline + intersection & Co-Section (ours) &\\
		\includegraphics[trim=0 10 59 20,clip,width=\linewidth]{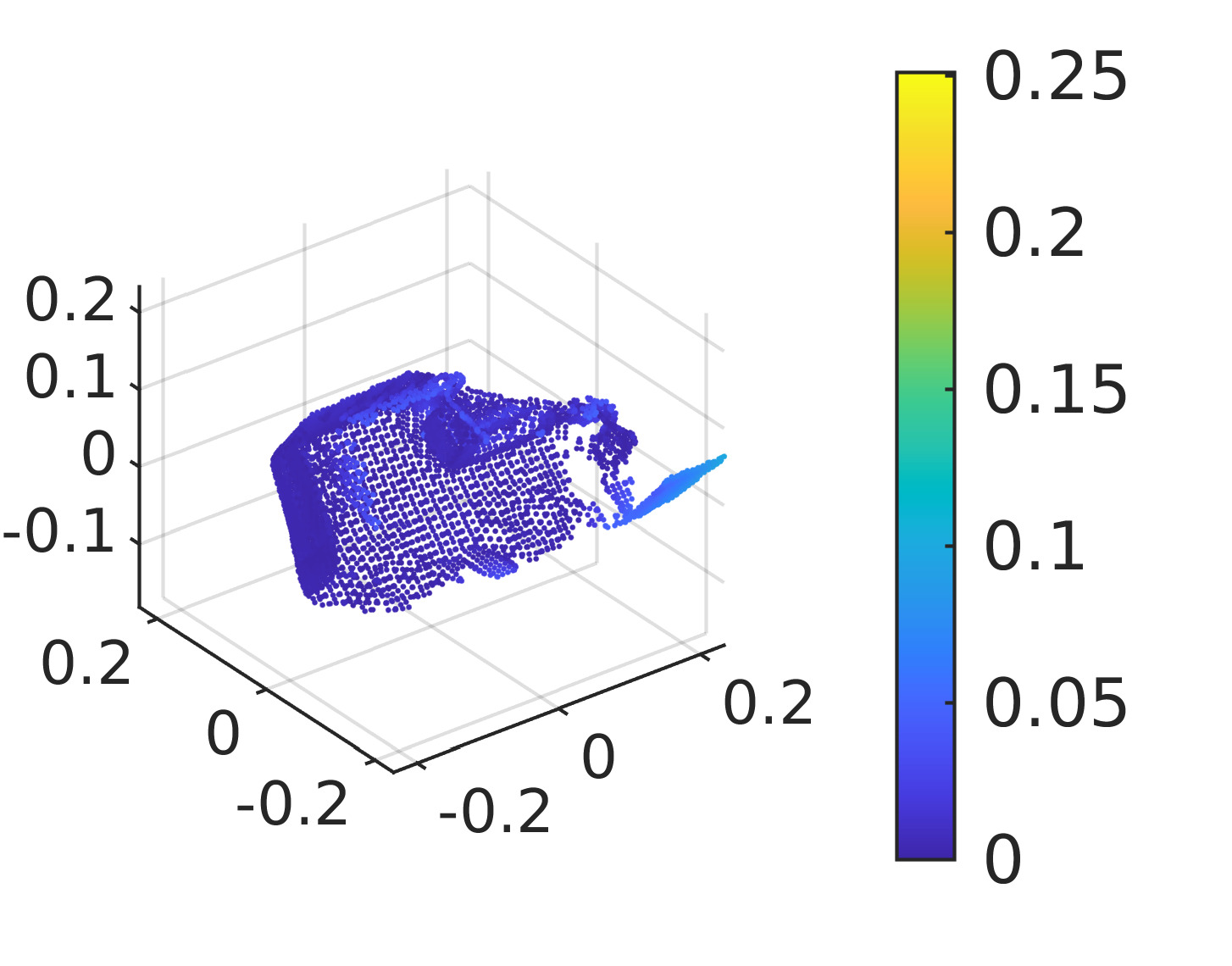} & \includegraphics[trim=0 10 59 20,clip,width=\linewidth]{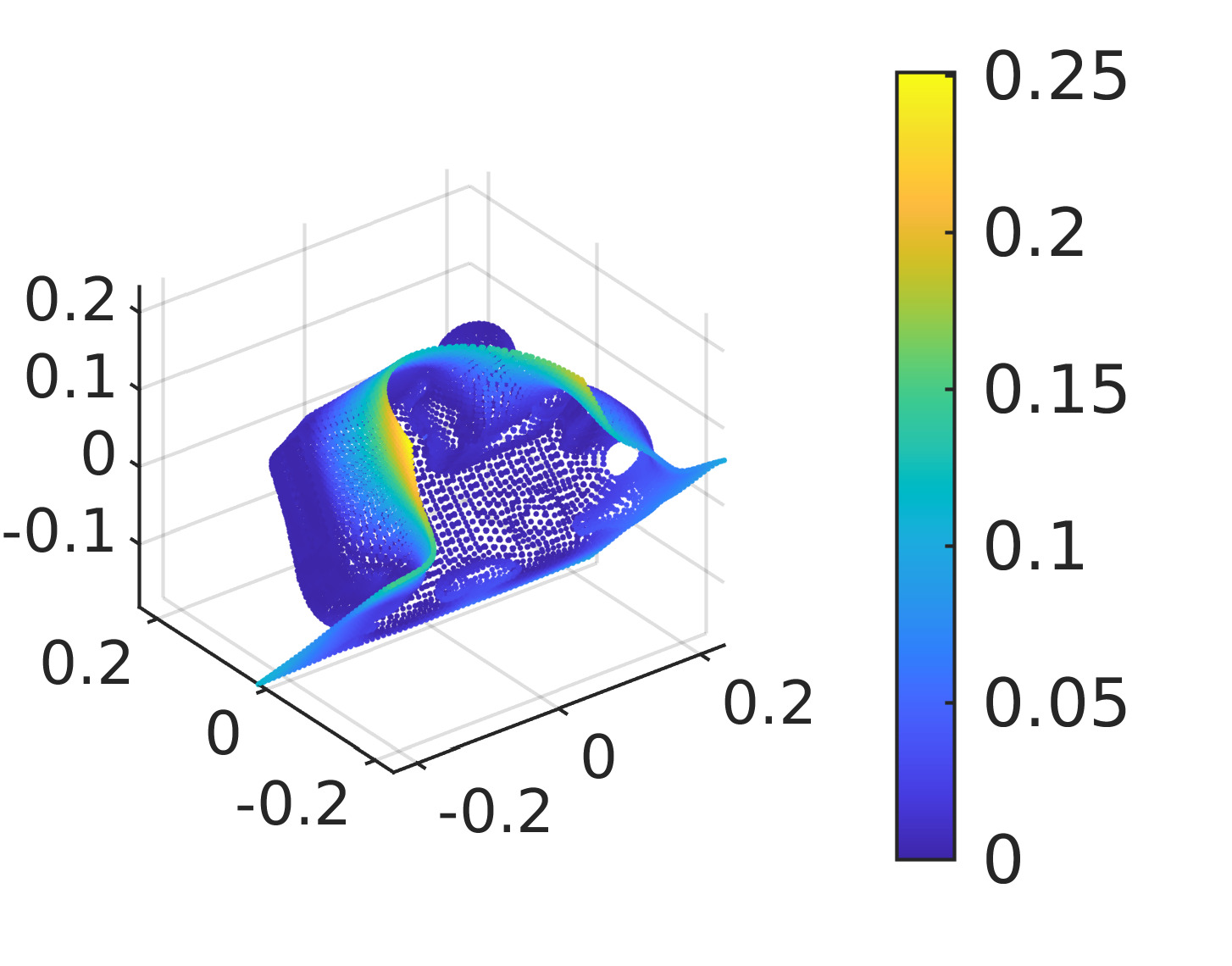} & \includegraphics[trim=0 10 59 20,clip,width=\linewidth]{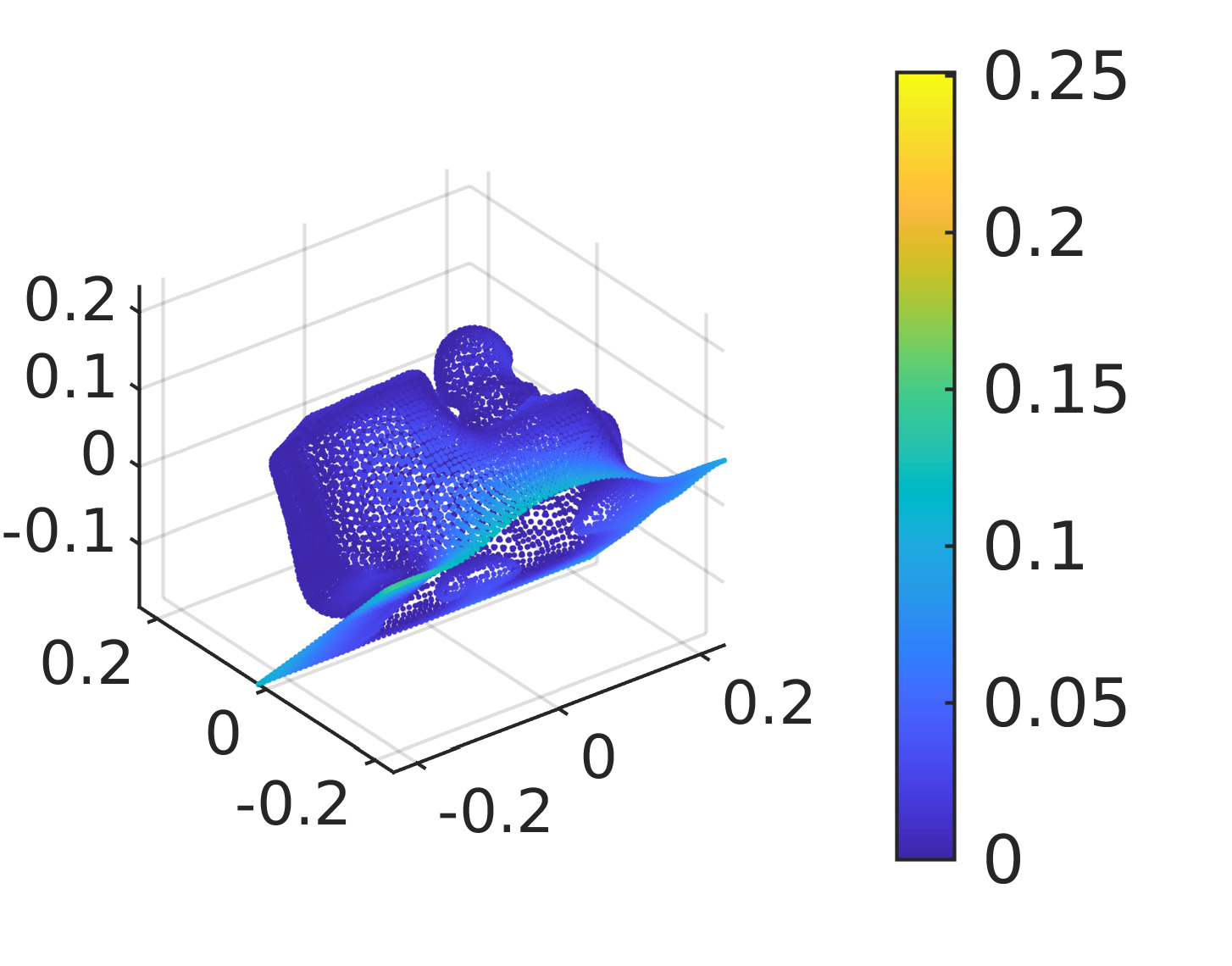} & \includegraphics[trim=0 10 59 20,clip,width=\linewidth]{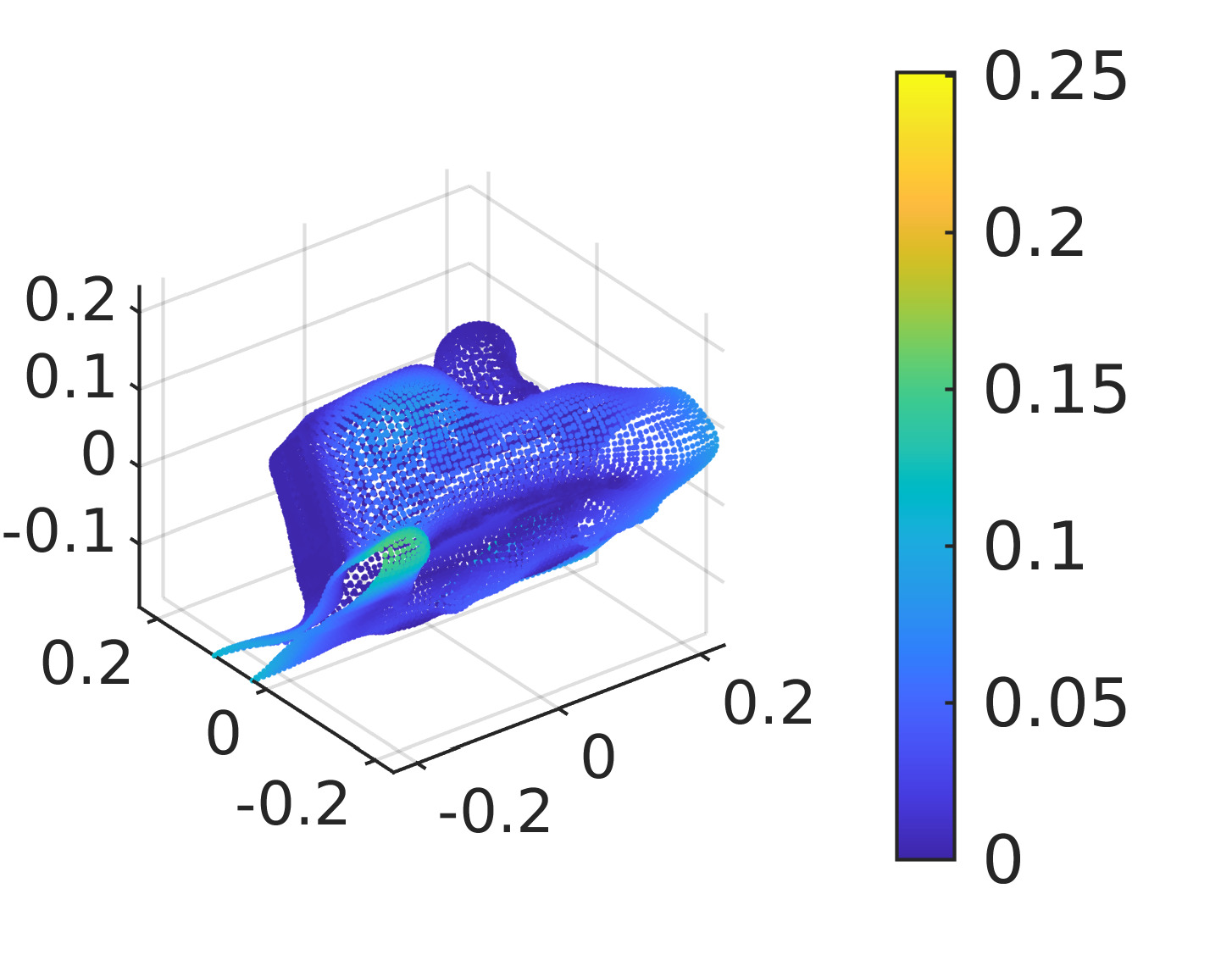} & \includegraphics[trim=0 10 59 20,clip,width=\linewidth]{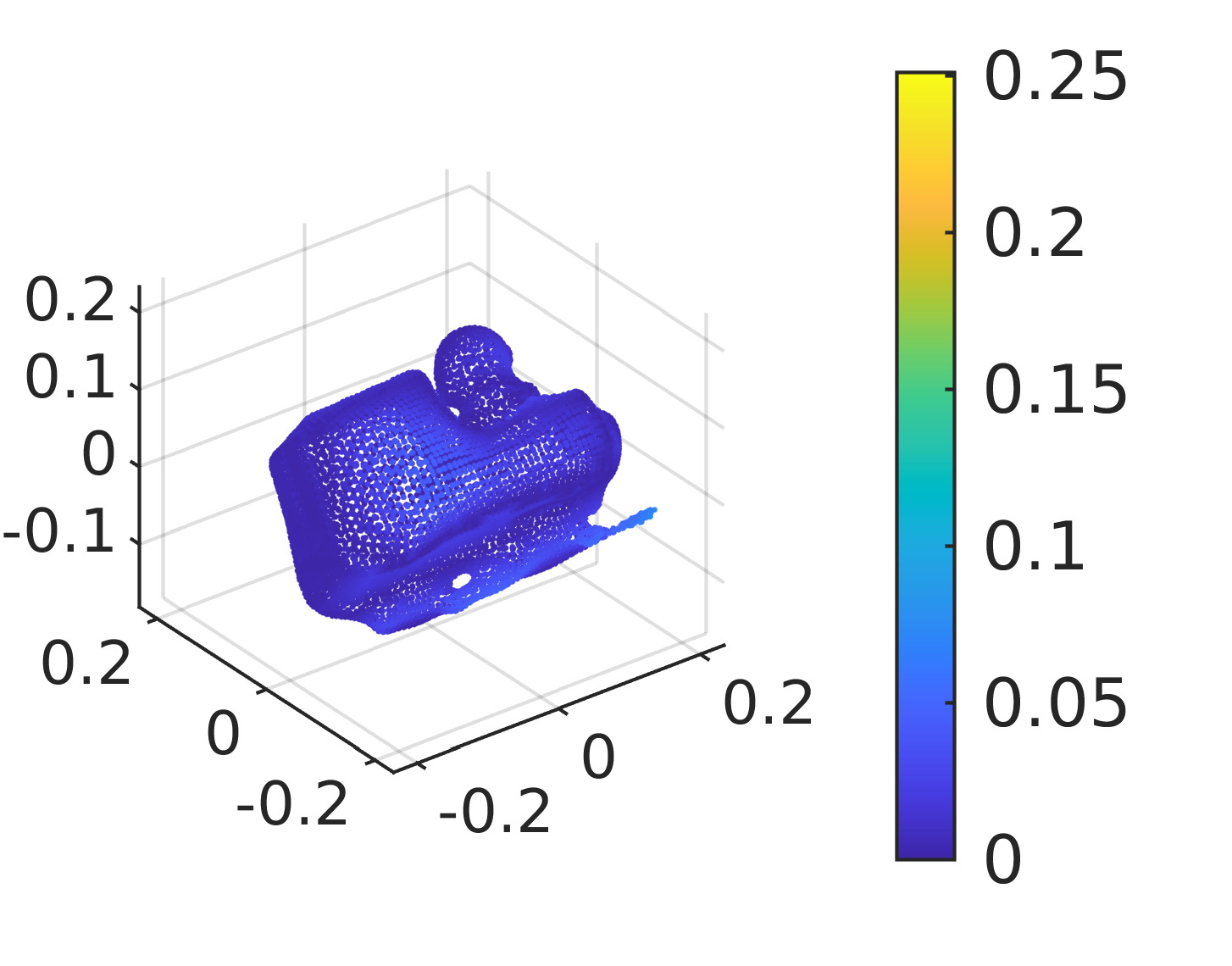} & \includegraphics[trim=122 0 13 0,clip,width=\linewidth]{figures/num_eval/acc_truck_final}\\
		\includegraphics[trim=0 10 59 20,clip,width=\linewidth]{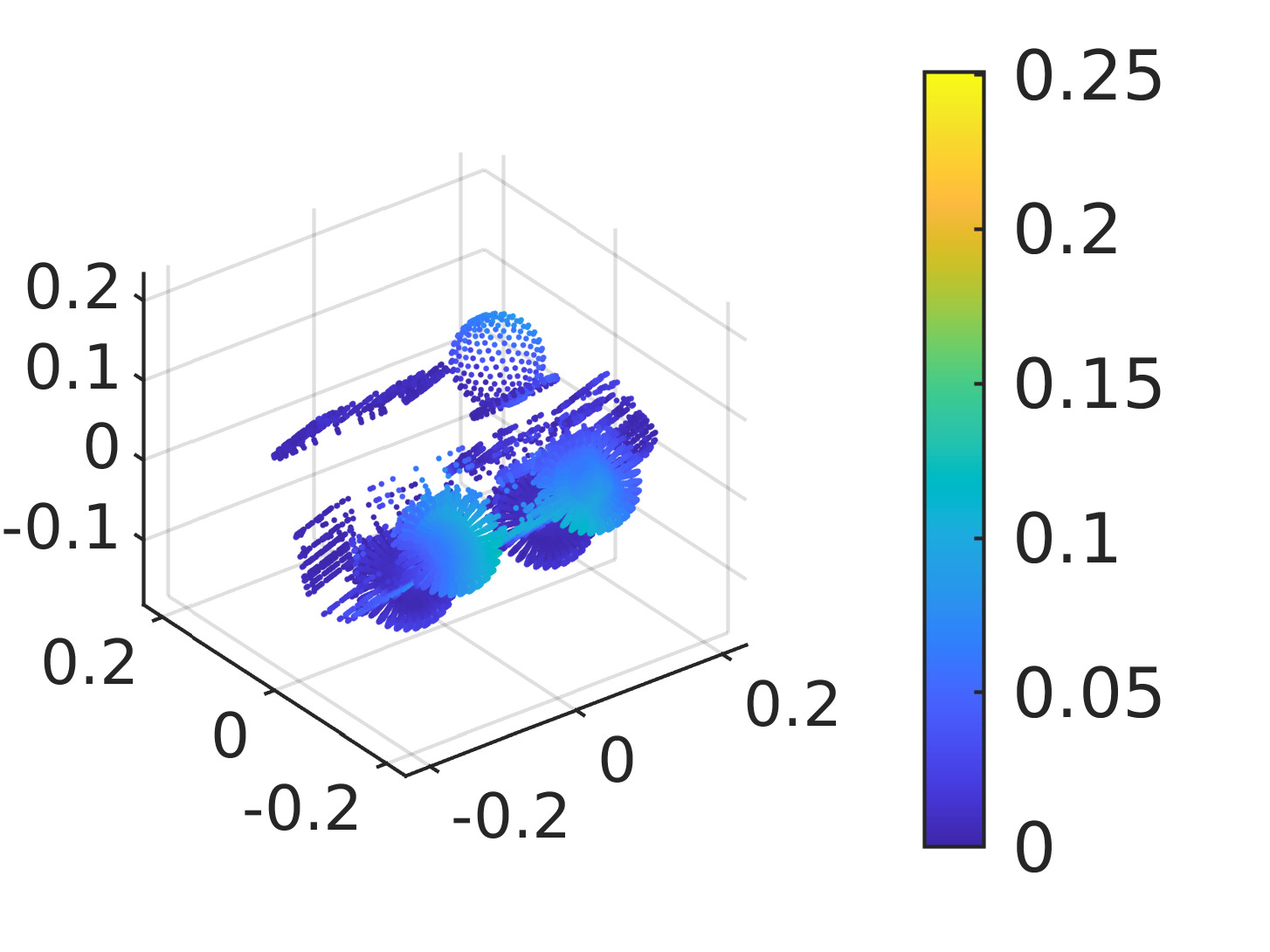} & \includegraphics[trim=0 10 59 20,clip,width=\linewidth]{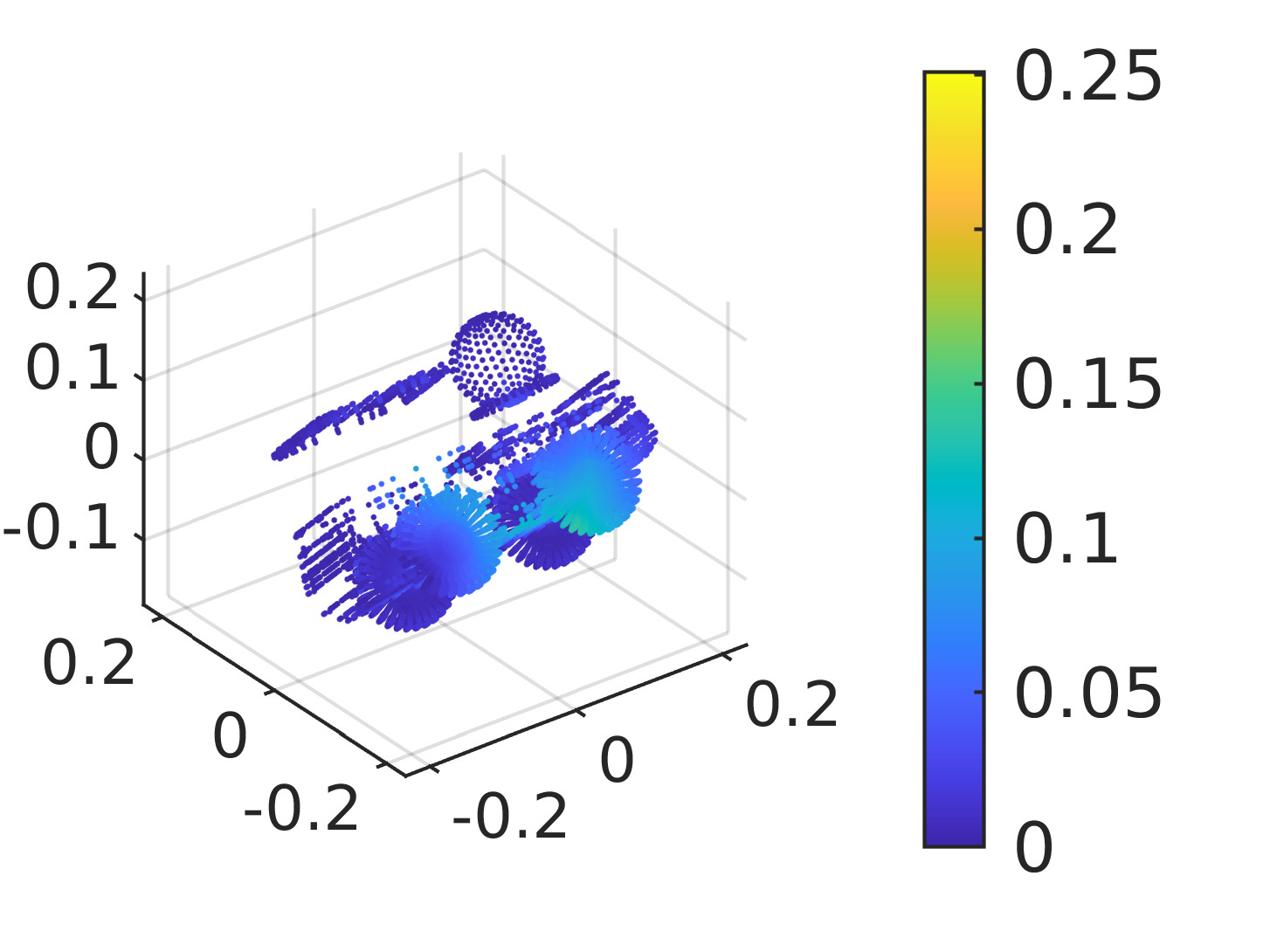} & \includegraphics[trim=0 10 59 20,clip,width=\linewidth]{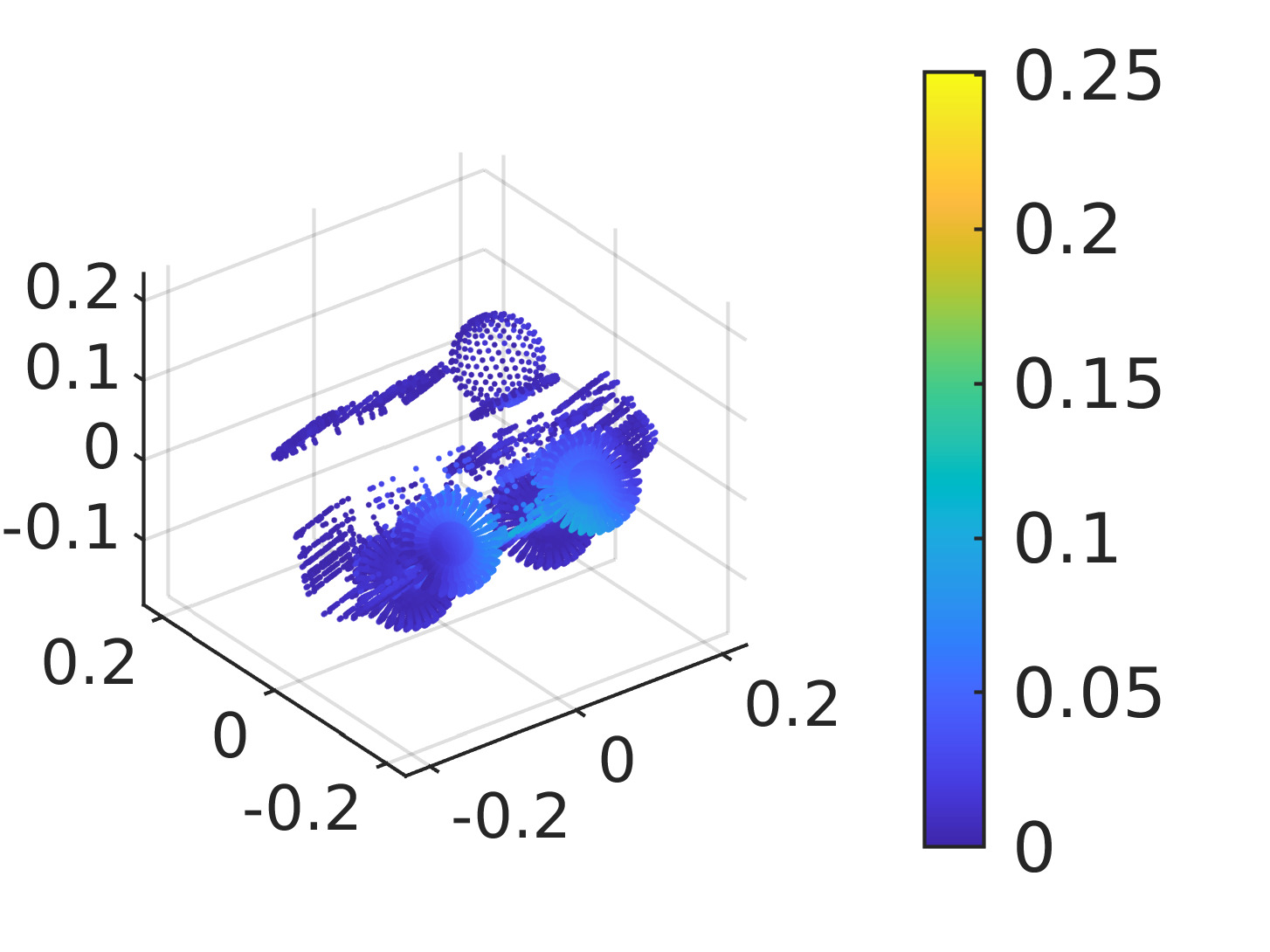} & \includegraphics[trim=0 10 59 20,clip,width=\linewidth]{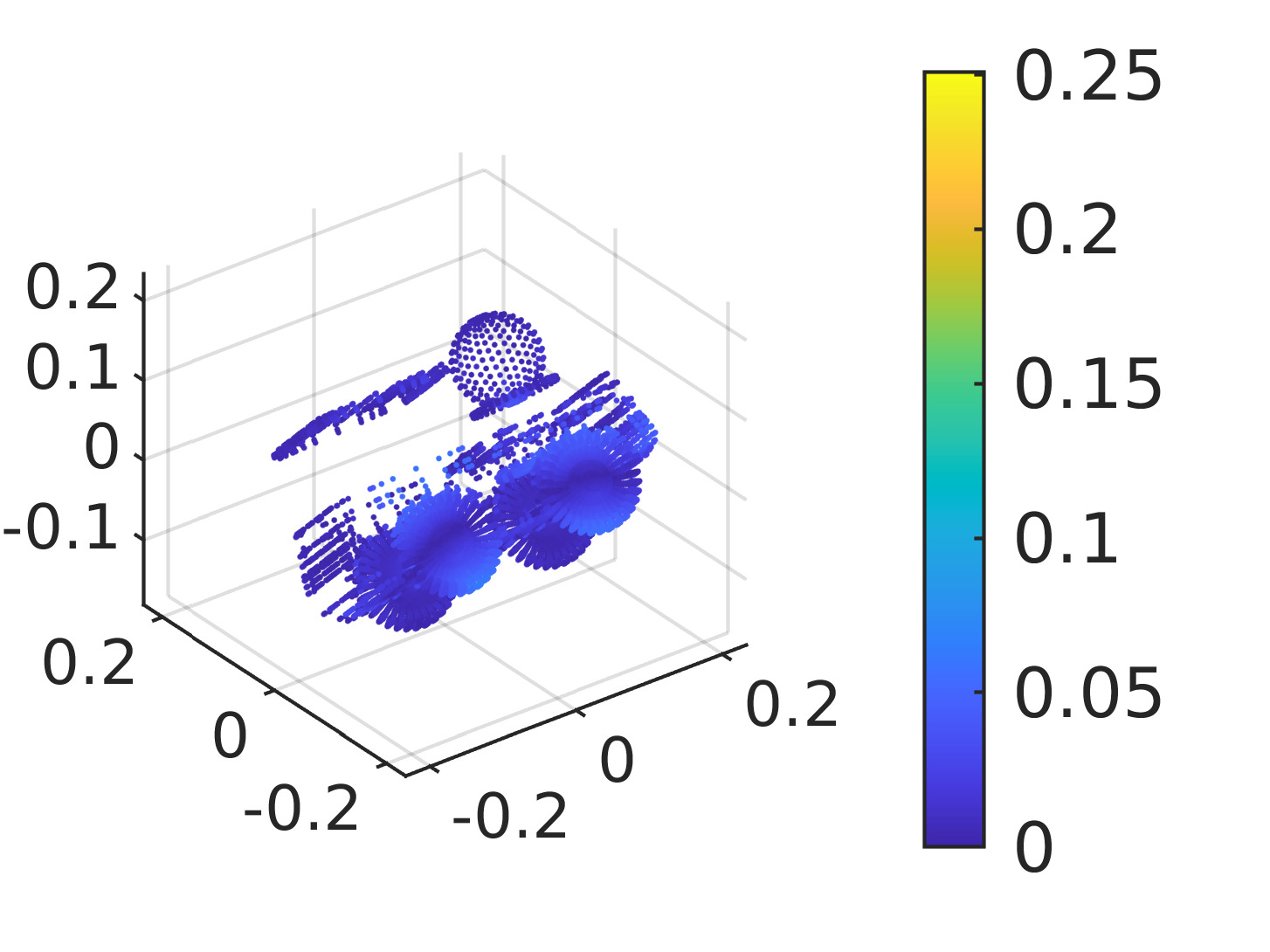} & \includegraphics[trim=0 10 59 20,clip,width=\linewidth]{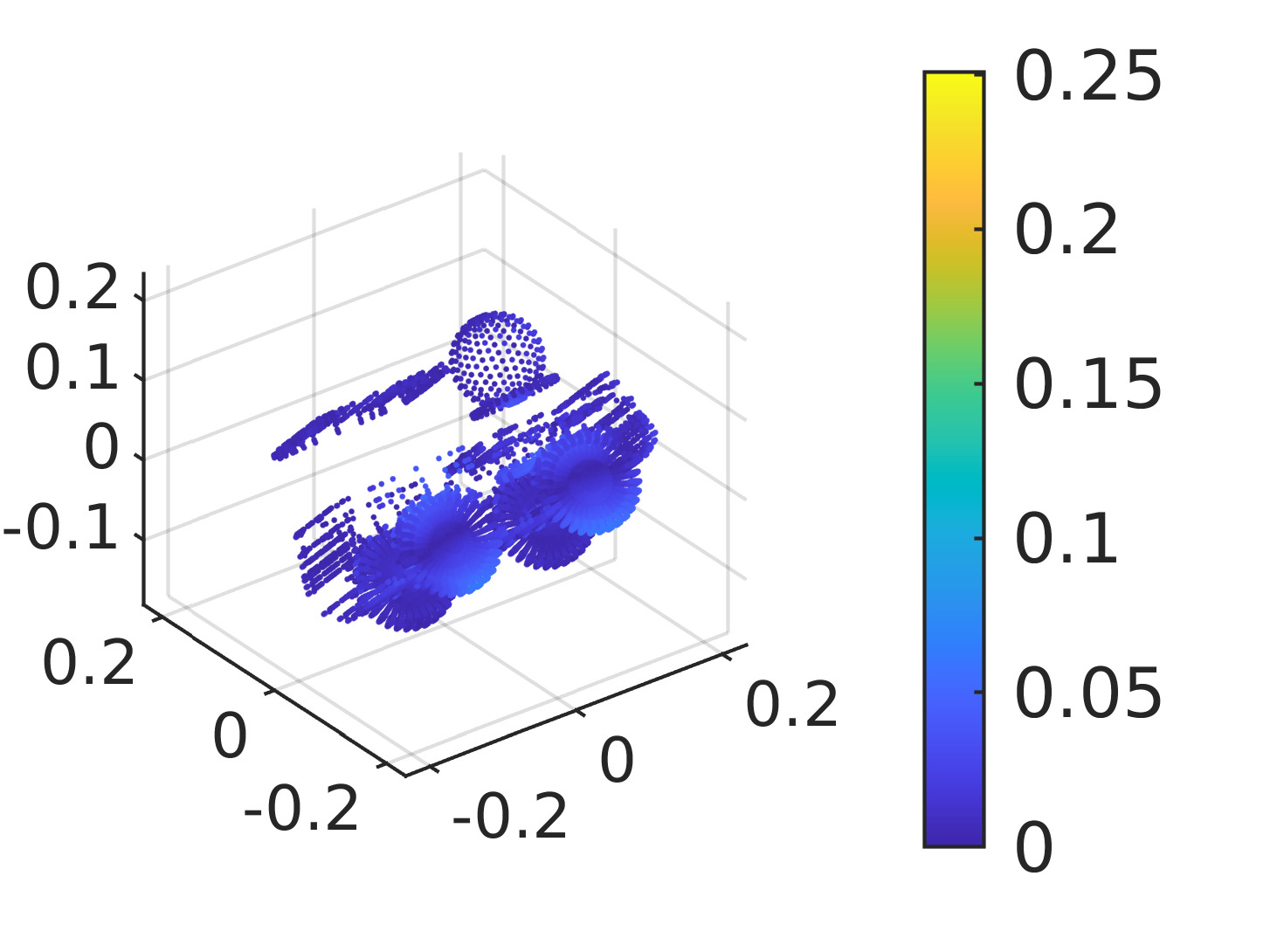} & \includegraphics[trim=122 0 13 0,clip,width=\linewidth]{figures/num_eval/comp_truck_final}\\
		\includegraphics[trim=0 10 59 20,clip,width=\linewidth]{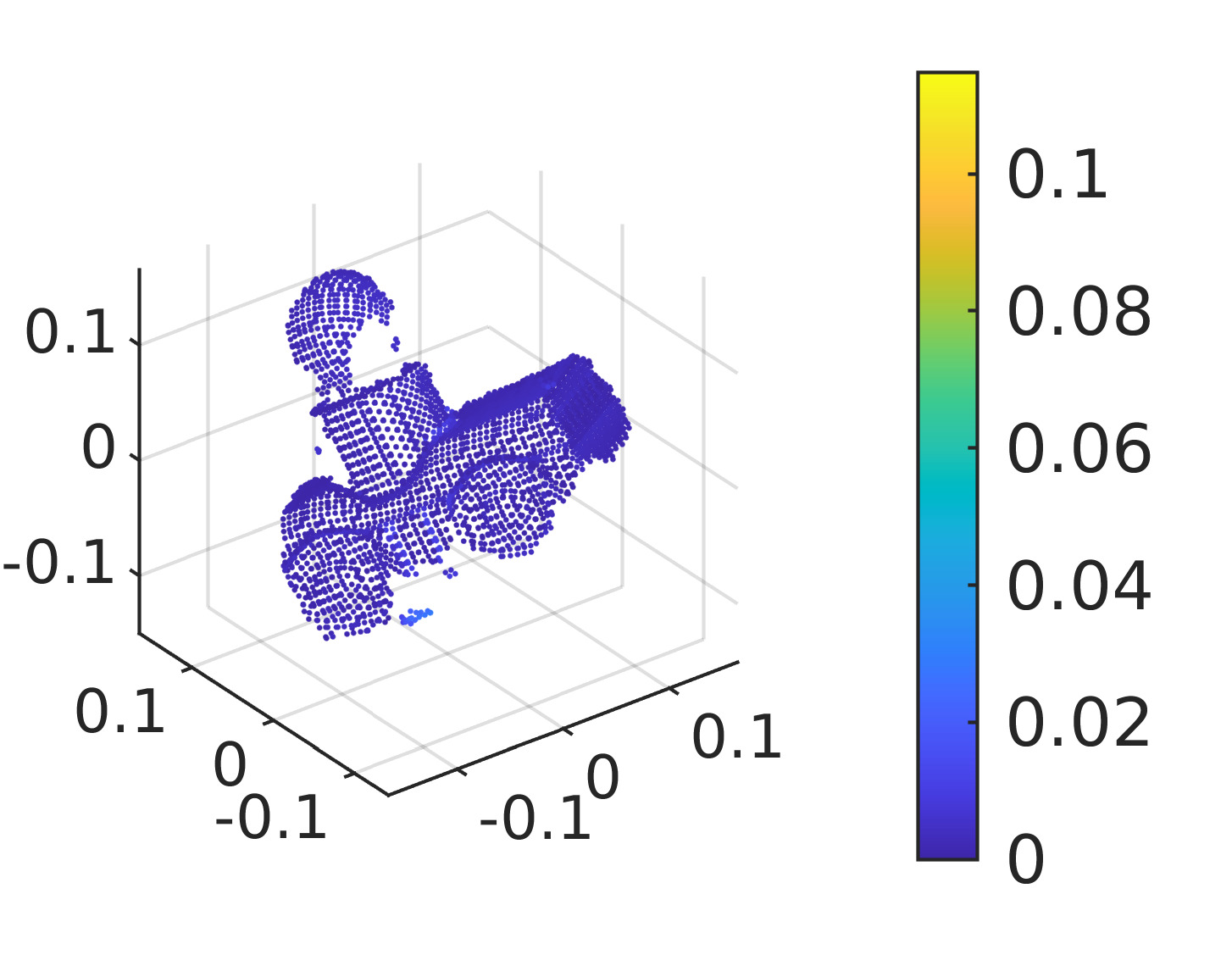} & \includegraphics[trim=0 10 59 20,clip,width=\linewidth]{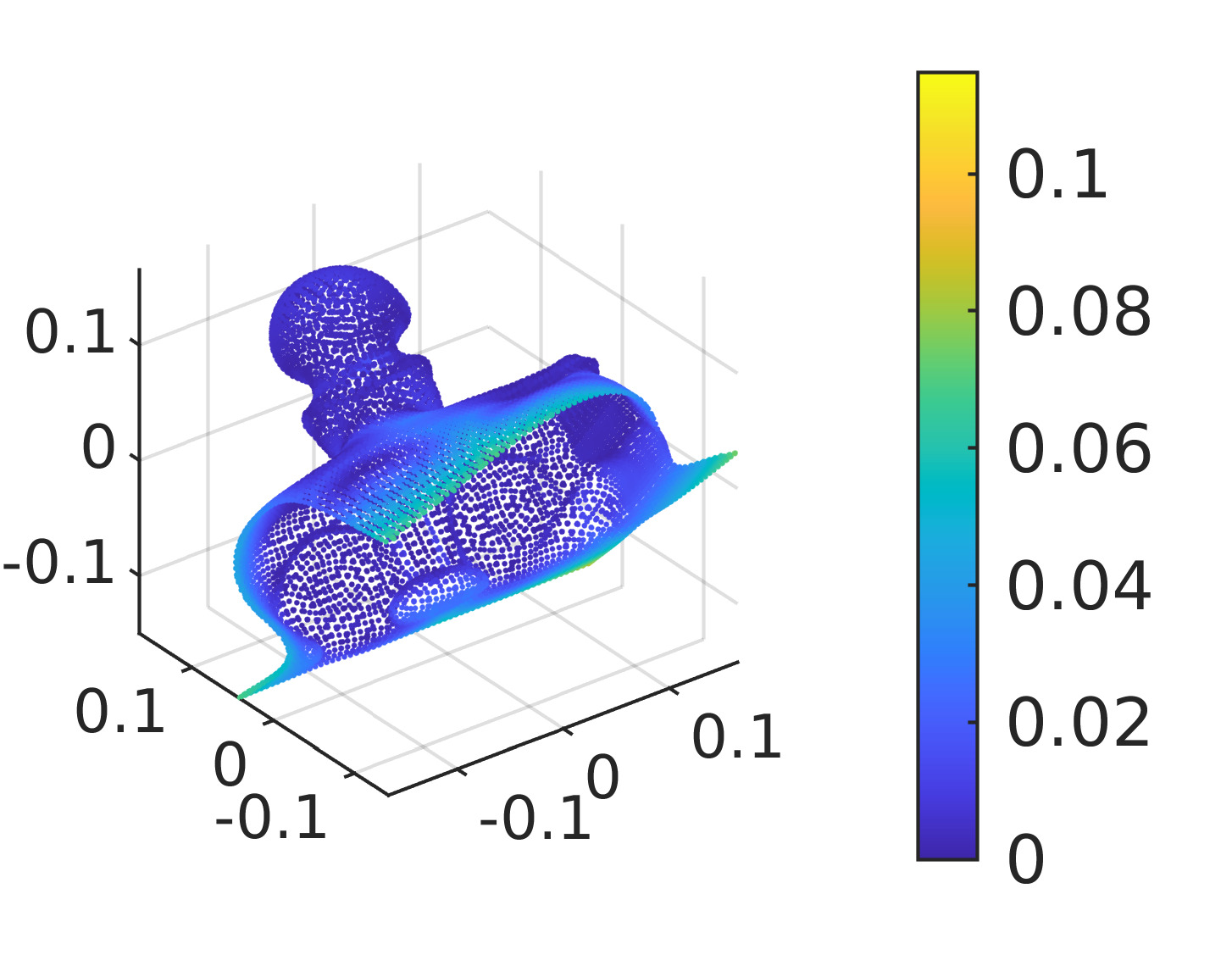} & \includegraphics[trim=0 10 59 20,clip,width=\linewidth]{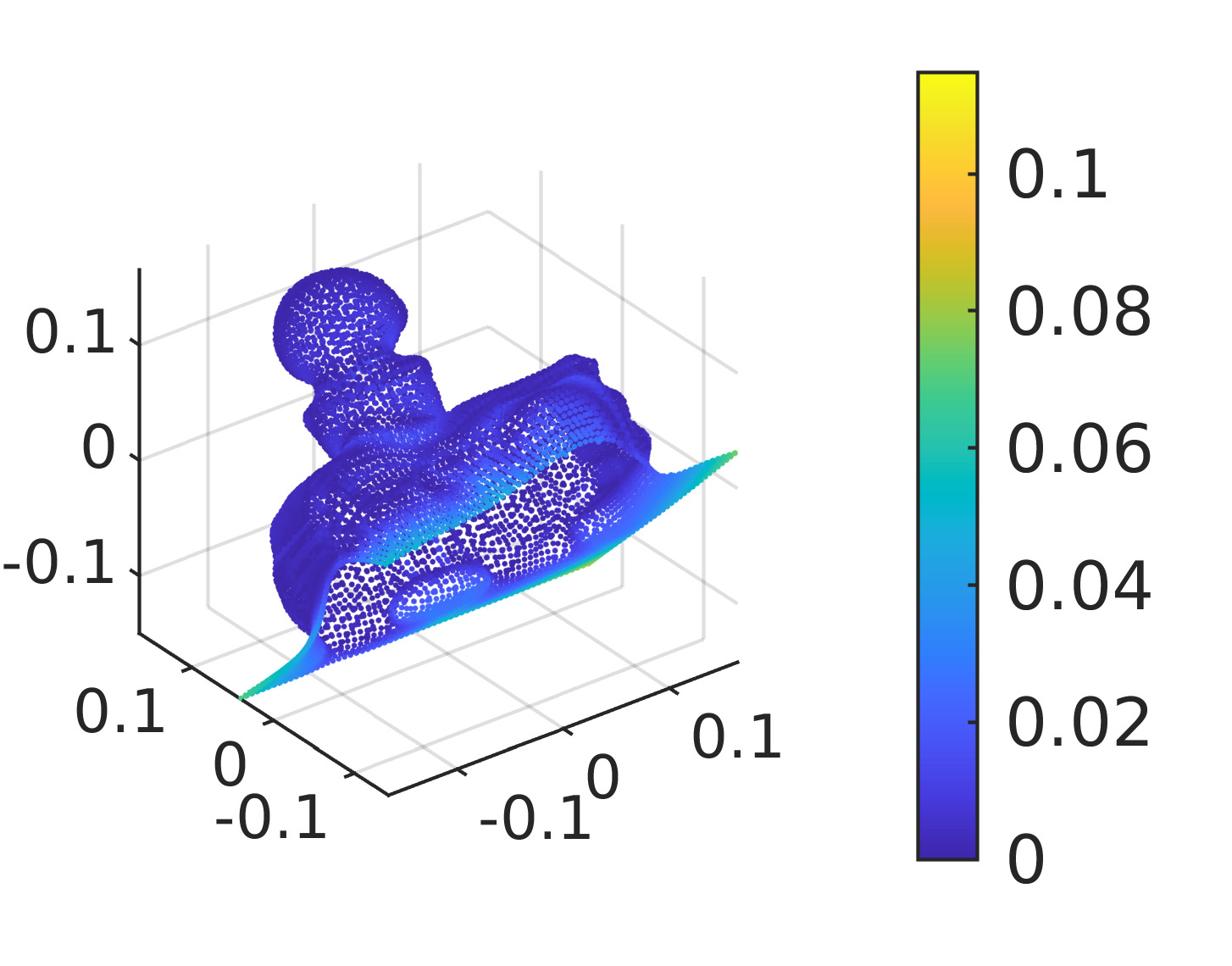} & \includegraphics[trim=0 10 59 20,clip,width=\linewidth]{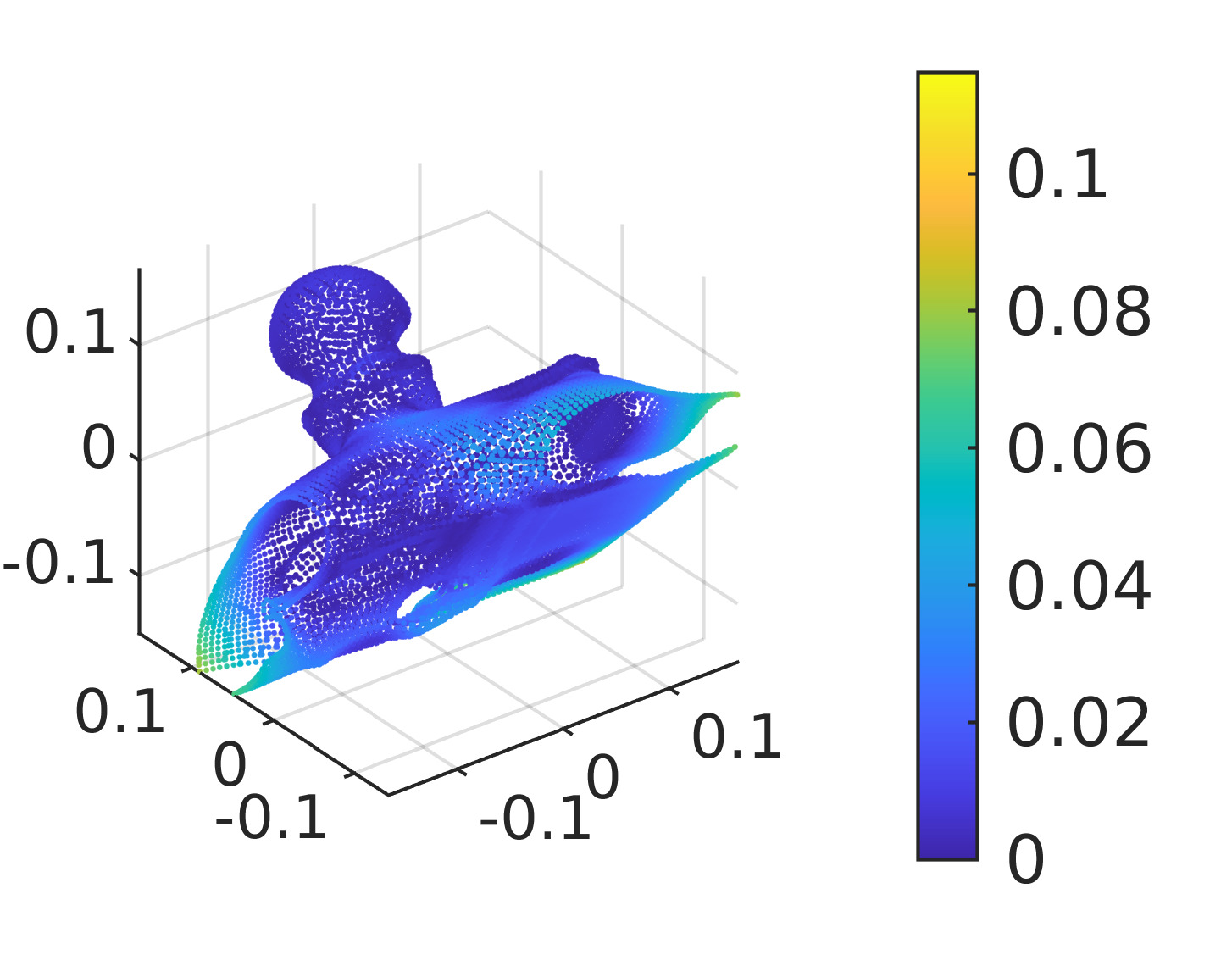} & \includegraphics[trim=0 10 59 20,clip,width=\linewidth]{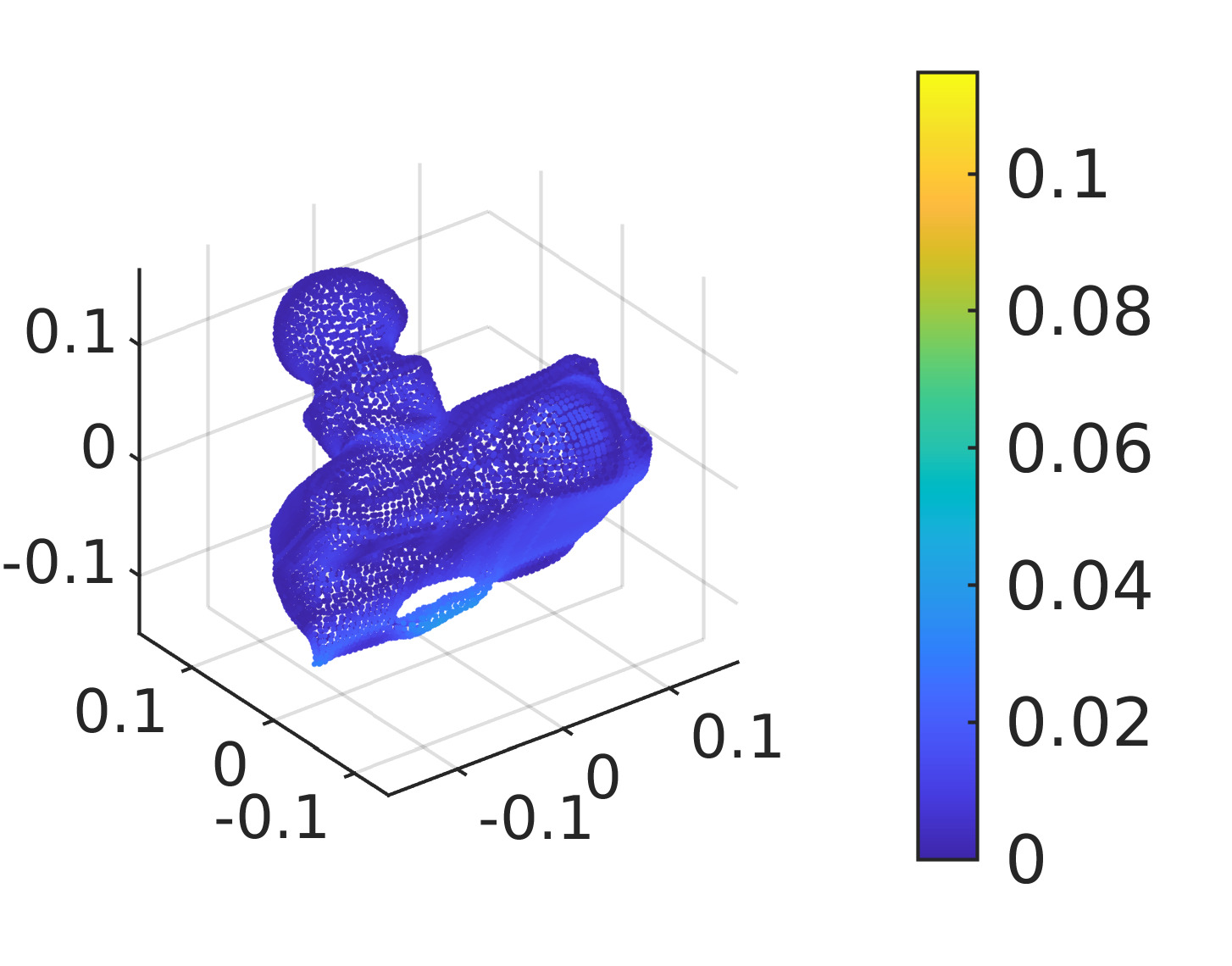} & \includegraphics[trim=125 0 10 0,clip,width=\linewidth]{figures/num_eval/acc_car_final}\\
		\includegraphics[trim=0 10 59 20,clip,width=\linewidth]{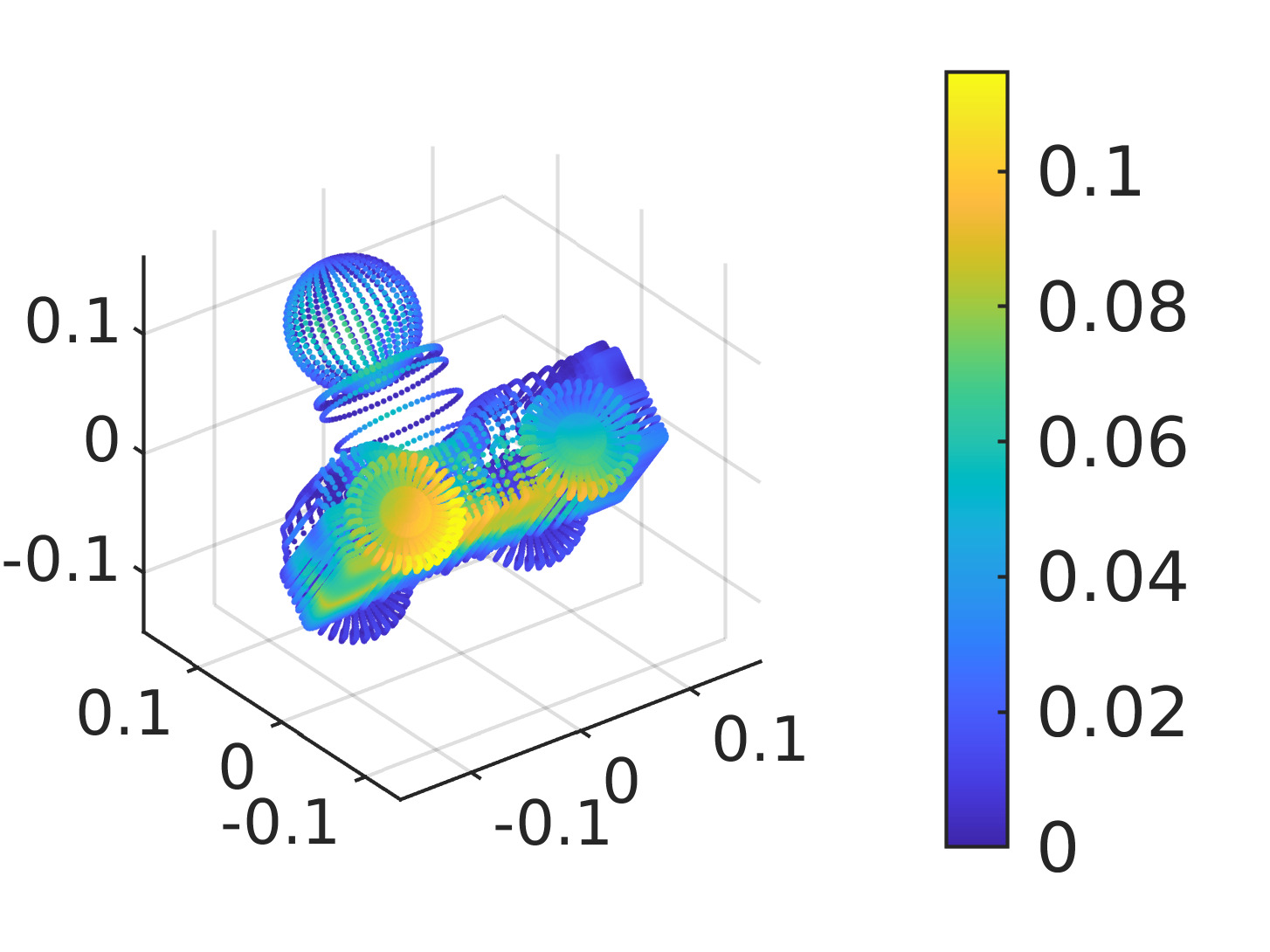} & \includegraphics[trim=0 10 59 20,clip,width=\linewidth]{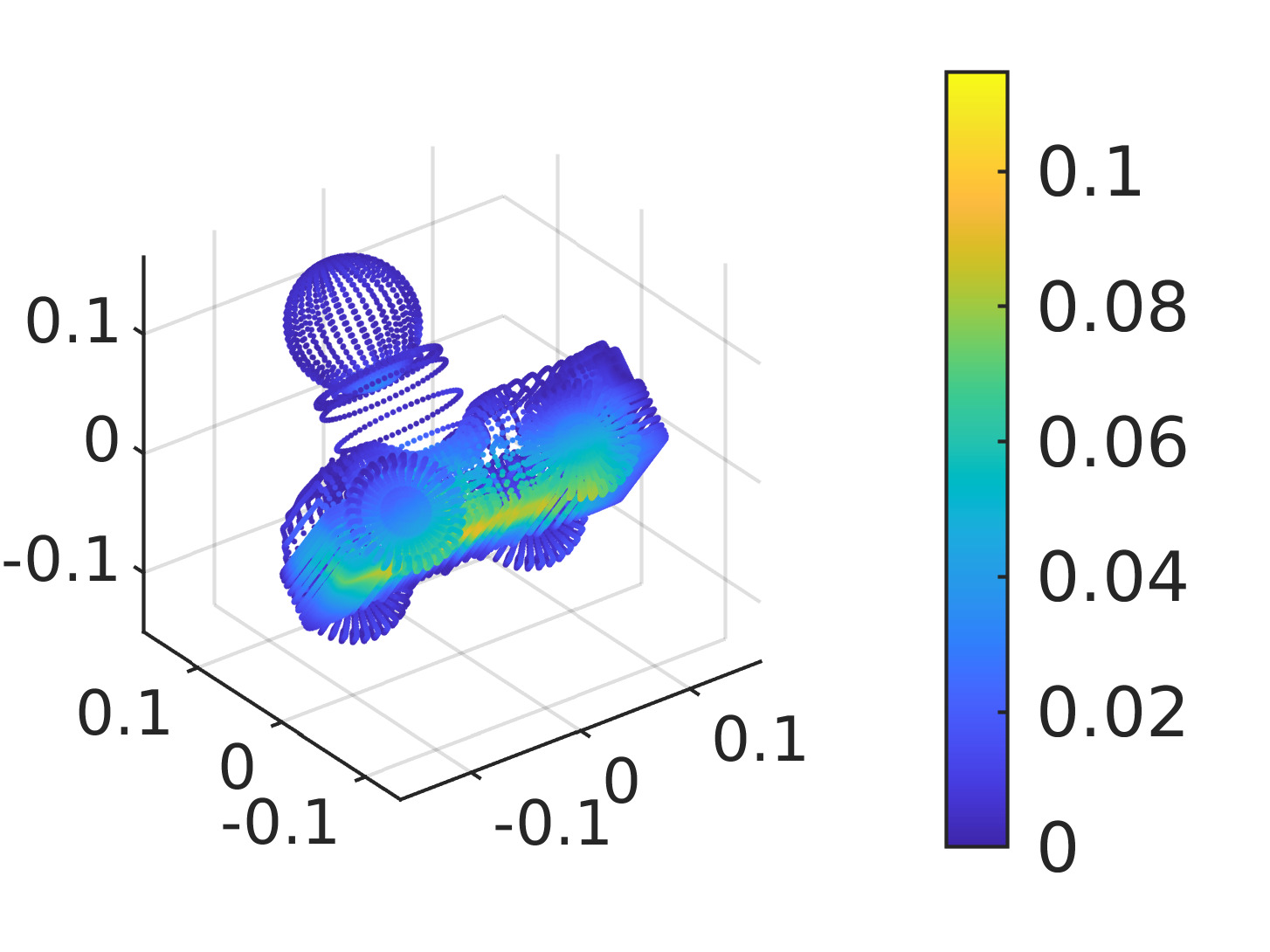} & \includegraphics[trim=0 10 59 20,clip,width=\linewidth]{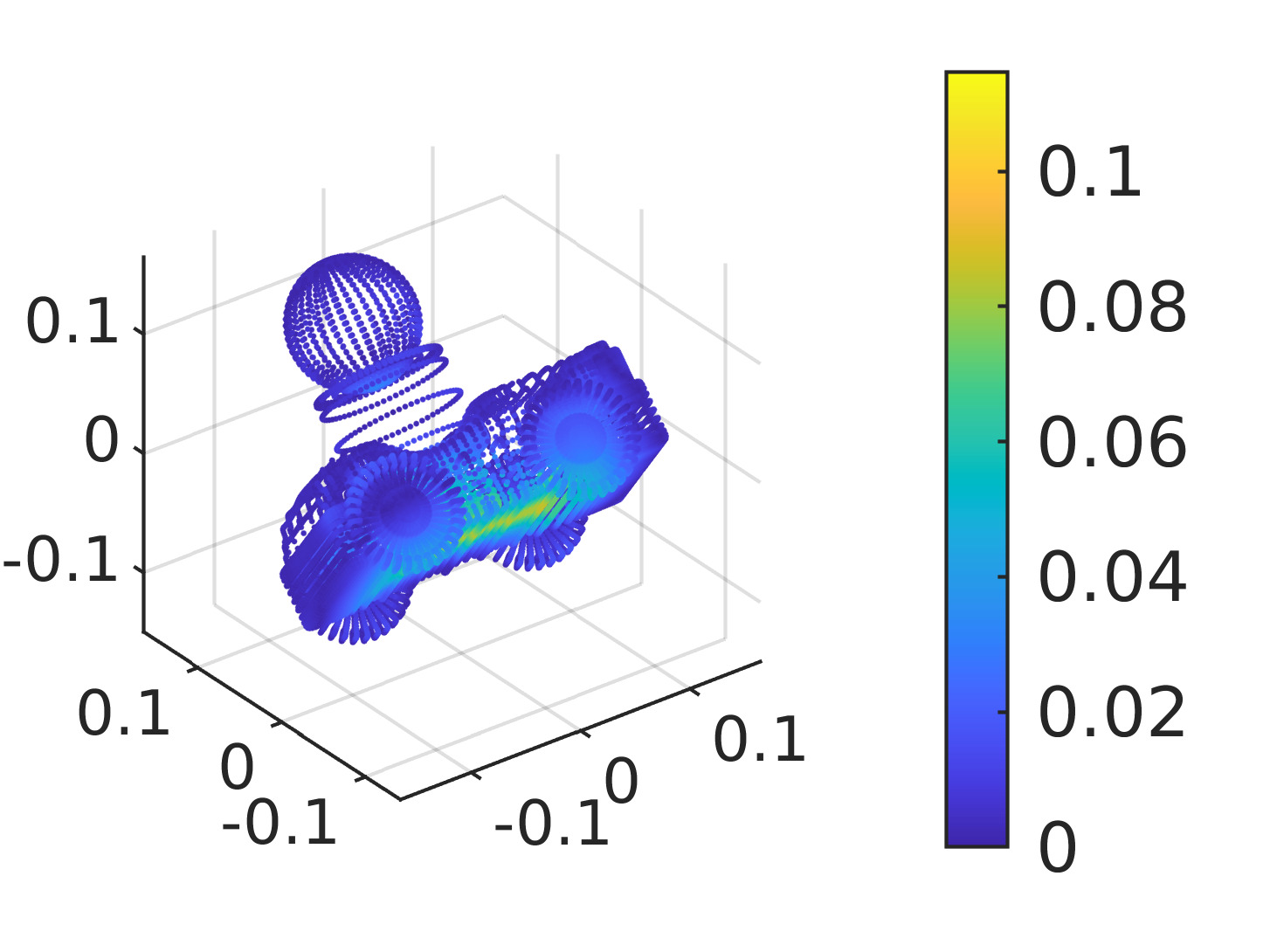} & \includegraphics[trim=0 10 59 20,clip,width=\linewidth]{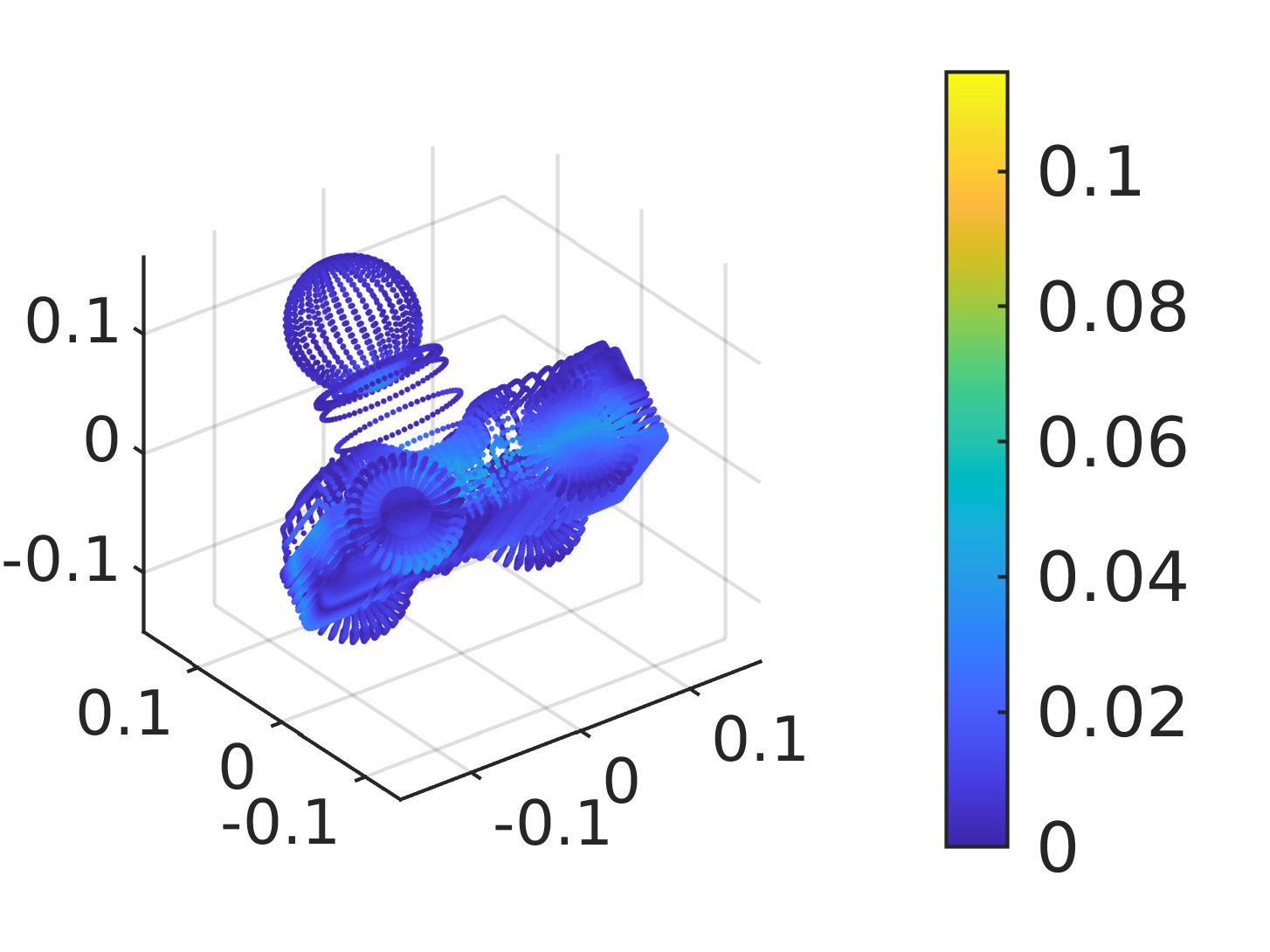} & \includegraphics[trim=0 10 59 20,clip,width=\linewidth]{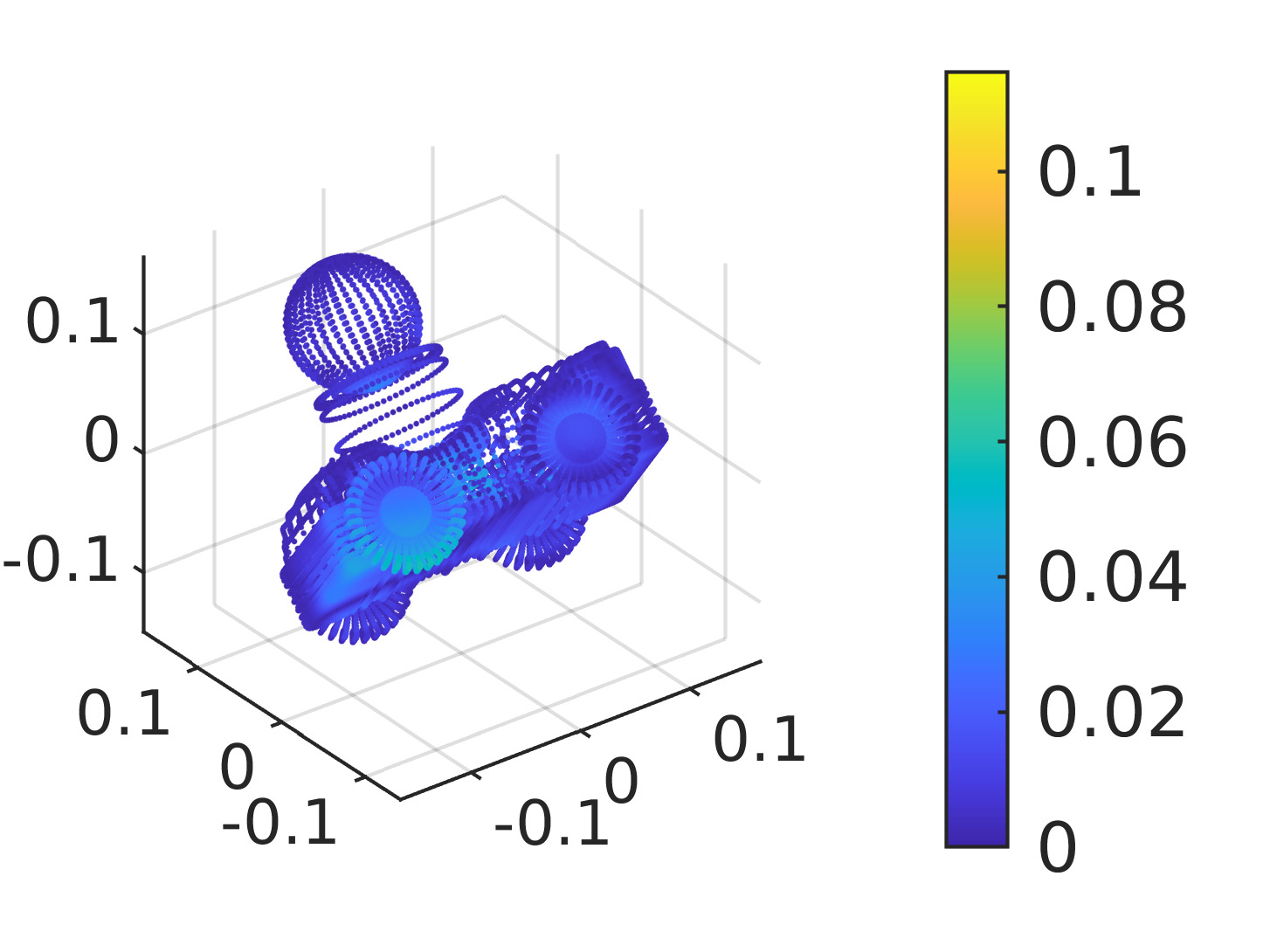} & \includegraphics[trim=125 0 10 0,clip,width=\linewidth]{figures/num_eval/comp_car_final}\\
	\end{tabular}
	\end{center}
	\caption{Accuracy (top row per object) and completeness (bottom row per object) of object shape reconstruction results on the Co-Fusion car4 sequence. Color indicates distance of mesh vertices to the other mesh. From top to bottom: truck, car. Our full approach is clearly superior to all other variants in accuracy as well as completeness.}
	\label{fig:quanrecon}
\end{figure*}

\textbf{Results.}
Table~\ref{tab:ablation} lists accuracy and completeness for the objects on the synthetic car4 sequence of the Co-Fusion dataset.
Our full optimization approach strongly improves the completeness of the shape towards the TSDF map.
For two of three objects, accuracy is reduced towards the TSDF map.
However, this is expected since our method inpaints new object parts which are not modelled by the TSDF.
Still, the accuracy of our full approach is better than any variant of the optimization method.
We observe that while the hull constraint tends to improve accuracy, the intersection constraint tends to improve completeness.
We conclude that both the hull and intersection constraints are important to achieve the performance of the full model.
In Fig.~\ref{fig:quanrecon} we show accuracy and completeness per mesh vertex (point-to-triangle distance).

Our method relies on accurate pose estimates by the underlying dynamic SLAM method.
In case of inaccurate pose estimates, hull and intersection volumes might become inaccurate as well, leading to inaccuracies in the surface reconstruction.
While we have not observed this problem in our experiments, a stronger weighting of the dataterm $E_\text{data}$ \eqref{eq:data_energy} could alleviate this problem to some extent.

\begin{table*}
	\footnotesize
	\begin{center}
	\begin{tabular}{lcccccc}
		\toprule
		       & \multicolumn{2}{c}{Truck} & \multicolumn{2}{c}{Car} & \multicolumn{2}{c}{Airplane}\\
	  Method & Acc & Comp & Acc & Comp & Acc & Comp\\
		\midrule
		TSDF & 0.00949544 & 0.0375519 & \bf 0.00204528 & 0.0277146 & \bf 0.0364824 & 0.0281357 \\
		baseline & 0.0333811 & 0.0327722 & 0.0154461 & 0.0196067 & 0.156872 & 0.0229247\\
		baseline + hull & 0.0234413 & 0.023931 & 0.0117728 & 0.0122979 & 0.150925 & 0.0174758\\
		baseline + intersection & 0.0325838 & 0.0116423 & 0.0163322 & 0.0109024 & 0.173195 & 0.0202907\\
		Co-Section (ours) & \bf 0.00878659 & \bf 0.0109651 & 0.00521935 & \bf 0.0102169 & 0.0871882 & \bf 0.016172\\
		\bottomrule
	\end{tabular}
	\end{center}
	\caption{Accuracy and completeness (lower is better) on the Co-Fusion car4 sequence for different variants of our method. Best in bold. Our full approach is clearly superior in completeness and also performs well in accuracy despite the fact that it ``guesses'' parts of the object surface from the constraints.}
	\label{tab:ablation}
\end{table*}

\textbf{Computation times.}
The intersection constraint can be computed for several voxels in parallel on a GPU.
In our experiments this computation takes under 10ms on average per volume.
Similarly, the data measurements and the hull constraint can be entered fast on parallel hardware.
The optimization itself is more time consuming.
For the object volumes in our experiments, our implementation of our variant of the Hessian-IMLS \cite{Schroers_2014_variational} achieves runtimes of several seconds (avg.~0.98s, peak 19.8s).
The peak runtime typically occurs when a mostly empty volume needs to be filled from sparse measurements after initialization.
We consider further parallelization and improvement of the runtime efficiency of our approach as future work.
Schroers \etal \cite{Schroers_2014_variational} report runtimes between 1s and 4s for volumes with resolutions up to $400^3$ (our resolution $64^3/256^3$ for objects/background, respectively).

\section{Conclusion}
In this paper we propose a novel energy minimization approach for object shape completion in dynamic scenes.
We incorporate hull and intersection constraints between objects into a formulation which optimizes for the implicit surface in a volumetric representation.
The data terms of our method are obtained with a dynamic object-level SLAM frontend which detects, segments, tracks and maps the objects in local TSDF volumes.
We demonstrate in our experiments that our formulation can achieve object shape completion which is physically plausible.
We analyze the contributions of the constraints for accuracy and completeness of object shape reconstruction.
Our approach improves completeness over the TSDF reconstruction and can achieve high accuracy even if parts of the object are unobserved.

In future work, we would like to investigate the incorporation of further regularization constraints to achieve improved scene reconstruction.
Currently, our method is not real-time capable. Optimization of a volume takes from under a second to several seconds.
This can still be interesting for backend optimization in a parallel thread with the frontend.
Devising methods for increasing the run-time efficiency is an interesting direction for future research.

\paragraph{\large Acknowledgements.}
We acknowledge support from Cyber Valley, the Max Planck Society, and the German Federal Ministry of Education and Research (BMBF) through the Tuebingen AI Center (FKZ: 01IS18039B).
The authors thank the International Max Planck Research School for Intelligent Systems (IMPRS-IS) for supporting Michael Strecke.

{\small
\bibliographystyle{ieee_fullname}
\bibliography{references}
}

\end{document}